\pdfoutput=1
\documentclass[11pt]{article}
\usepackage{ACL2023}
\usepackage{times}
\usepackage{latexsym}
\usepackage[T1]{fontenc}
\usepackage[utf8]{inputenc}
\usepackage{microtype}
\usepackage{inconsolata}
\usepackage{linguex}
\usepackage{graphicx}
\usepackage{booktabs}
\usepackage[all]{nowidow}
\usepackage{enumerate}

\title{How poor is the stimulus? Evaluating hierarchical generalization in neural networks trained on child-directed speech}
\author{Aditya Yedetore$^{*1}$, Tal Linzen$^2$, Robert Frank$^3$, R. Thomas McCoy$^{*4}$\\ $^1$Boston University, $^2$New York University, $^3$Yale University, $^4$Princeton University \\ \texttt{yedetore@bu.edu},  \texttt{linzen@nyu.edu}, \texttt{robert.frank@yale.edu}, \\\texttt{tom.mccoy@princeton.edu}
}

\begin{document}

\renewcommand{\firstrefdash}{}

\maketitle
\begin{abstract}
When acquiring syntax, children consistently choose hierarchical rules over competing non-hierarchical possibilities. Is this preference due to a learning bias for hierarchical structure, or due to more general biases that interact with hierarchical cues in children's linguistic input? We explore these possibilities by training LSTMs and Transformers---two types of neural networks without a hierarchical bias---on data similar in quantity and content to children's linguistic input: text from the CHILDES corpus. We then evaluate what these models have learned about English yes/no questions, a phenomenon for which hierarchical structure is crucial. We find that, though they perform well at capturing the surface statistics of child-directed speech (as measured by perplexity), both model types generalize in a way more consistent with an incorrect linear rule than the correct hierarchical rule. These results suggest that human-like generalization from text alone requires stronger biases than the general sequence-processing biases of standard neural network architectures.
\end{abstract}

\newcommand\blfootnote[1]{
  \begingroup
  \renewcommand\thefootnote{}\footnote{#1}
  \addtocounter{footnote}{-1}
  \endgroup
}
\blfootnote{$^*$ Work done while at Johns Hopkins University.}

\section{Introduction}\label{sec:intro}

Syntax is driven by hierarchical structure, yet we typically encounter sentences as linear sequences of words.
How do children come to recognize the hierarchical nature of the languages they acquire?
Some argue that humans must have a hierarchical inductive bias---an innate predisposition for hierarchical structure \cite{chomsky1965aspects,chomsky1980rules}.
An alternative view \cite[e.g.,][]{lewis2001learnability} is that no such bias is necessary: there may be clear evidence for hierarchical structure in children's input, so that children would choose hierarchical rules even without a hierarchical bias. 

At first blush, recent work in natural language processing (NLP) may seem to indicate that no hierarchical bias is necessary. 
Neural networks trained on naturally-occurring text perform impressively on syntactic evaluations even though they have no explicit syntactic structure built into them \cite[e.g.,][]{gulordava2018colorless,wilcox2018rnn,warstadt2020blimp}. 
However, these results do not provide strong evidence about the learning biases required to learn language from the data available to humans because these models receive very different training data than humans do \cite{warstadt2022artificial}. 
First, NLP models are typically trained on far more data than children receive, so models have more opportunities to encounter rare syntactic structures \cite{linzen2020accelerate}.
Second, most training sets in NLP are built from Internet text (e.g., Wikipedia), which differs qualitatively from 
the utterances that children typically hear; e.g., sentences in Wikipedia are on average 25 words long  \cite{yasseri2012wikipedia}, 
compared to 5 words for sentences in the North American English subset of the CHILDES corpus of child-directed speech  \cite{macwhinney2000childes}.

In this work, to evaluate if neural networks without a hierarchical bias generalize like children do, we train models on text\footnote{Section \ref{sec:input} discusses other input types (e.g., visual input).} comparable to the sentences in children's linguistic input: English data from CHILDES. We then analyze what they have learned about the relationship between declarative sentences, such as \ref{ex:checkers_statement}, and their corresponding yes/no questions, such as \ref{ex:checkers_question}:

\setlength{\Exlabelwidth}{0.7em}
\setlength{\SubExleftmargin}{1.35em}
\setlength{\Extopsep}{6pt}

    \ex. \a. Those \underline{\textcolor{blue}{\textbf{are}}} your checkers.\label{ex:checkers_statement}
    \b. \underline{\textcolor{blue}{\textbf{Are}}} those your checkers?\label{ex:checkers_question}

Crucially, nearly all naturally-occurring yes/no questions are consistent with two rules: one based on hierarchical structure \ref{ex:hierarchicalq_rule}, and one based on linear order \ref{ex:linearq_rule}:\footnote{In past work these rules have been framed as transformations named \textsc{Move-First} and \textsc{Move-Main} \cite{mccoy2020trees}. We instead follow \citet{berwick2011revisited} and frame the child's knowledge as a relationship between sentences.}\textsuperscript{,}\footnote{Though these two rules are the most prominent in prior literature, other rules are possible; see Section \ref{sec:question_formation_results}.}

\ex. \textsc{HierarchicalQ}: The auxiliary at the start of a yes/no question corresponds to the \textbf{main} auxiliary of the corresponding declarative.\label{ex:hierarchicalq_rule} 

\ex. \textsc{LinearQ}:  The auxiliary at the start of a yes/no question corresponds to the \textbf{first} auxiliary of the corresponding declarative.\label{ex:linearq_rule}  

Despite the scarcity of evidence disambiguating these rules,  children reliably favor \textsc{HierarchicalQ} \cite{crain1987structure}, albeit with occasional errors consistent with \textsc{LinearQ} \cite{ambridge2008structure}. 
Yes/no questions thus are  a prime candidate for an aspect of English syntax for which human-like generalization requires a hierarchical bias. 
We evaluate yes/no question performance in LSTMs and Transformers, two neural-network architectures that have no inherent hierarchical inductive bias \cite{mccoy2020trees,petty2021transformers}. These architectures employ different computational mechanisms, so consistent results across both would indicate that our results are not due to  idiosyncrasies of one particular architecture.

To investigate if models generalize more consistently with the hierarchical or linear rule,
we evaluate them on cases where the rules make different predictions, such as  \ref{different_predictions}:
under \textsc{HierarchicalQ}, the question that corresponds to \ref{ex:declarative_from_cfg} is \ref{ex:hierarchical_q}, whereas under \textsc{LinearQ} it is \ref{ex:linear_q}.

\ex. \label{different_predictions}
\a. The boy who \textcolor{red}{\textbf{has}} talked \underline{\textcolor{blue}{\textbf{can}}} read.\label{ex:declarative_from_cfg}
\b. \underline{\color{blue}\textbf{Can}} \color{black} the boy who  \textcolor{red}{\textbf{has}} talked   \underline{\hspace{0.5cm}} read?\label{ex:hierarchical_q}
\c. *\color{red}\textbf{Has} \color{black} the boy who \underline{\hspace{0.5cm}} talked \underline{\color{blue}\textbf{can} \color{black}} read? \label{ex:linear_q}

\noindent

\noindent
We find that across several ways of framing the learning task, models fail to learn \textsc{HierarchicalQ}. Instead, they generalize in ways that depend on linear order and on the identities of specific words.
These results suggest that children's training data, if taken to be words alone, may not contain enough hierarchical cues to encourage hierarchical generalization in a learner without a hierarchical bias.
Thus, explaining human acquisition of syntax may require postulating that humans have stronger inductive biases than those of LSTMs and Transformers, or that information other than word sequences plays a crucial role.\footnote{GitHub repo with data and code: \href{https://github.com/adityayedetore/lm-povstim-with-childes}{\texttt{https://github.com/\\adityayedetore/lm-povstim-with-childes}}.}

\section{Background}
    Though \textsc{HierarchicalQ} and \textsc{LinearQ} often make the same predictions, the evidence in children's input may still favor \textsc{HierarchicalQ}.
    The most straightforward evidence would be utterances that directly disambiguate the rules, such as~\ref{ex:hierarchical_q}.  \citet{pullum2002empirical} show that disambiguating examples appear in the \textit{Wall Street Journal}, in literature, and arguably 
    in child-directed speech, but
    direct evidence may still be too rare to robustly support \textsc{HierarchicalQ} \cite{legate2002reassessment}.
    Nonetheless, children might conclude that yes/no questions obey \textsc{HierarchicalQ} rather than \textsc{LinearQ} based on \textit{indirect} evidence---evidence that \textit{other} syntactic phenomena are hierarchical \cite{mulligan2021structure}. 
    
    To test if the cues favoring \textsc{HierarchicalQ} render a hierarchical bias unnecessary, we study how well non-hierarchically-biased models acquire English yes/no questions.
    Several prior papers have used this approach, but their training data differed from children's input in important ways: some used synthetic datasets \cite{lewis2001learnability,frank2007transformational,clark2007polynomial,mccoy2020trees}, others used massive Internet corpora \cite{lin2019open,warstadt2020neural}, and those that used child-directed speech simplified the data by replacing each word with its part of speech \cite{perfors2011learnability,bod2012empiricist}. 
    We used training data closer to children's input, namely sentences from CHILDES with word identities preserved, rather than being converted to parts of speech. 
    Two other recent works have also trained neural networks on CHILDES data \cite{pannitto2020recurrent,huebner2021babyberta}, but neither investigated yes/no questions.
    
    One particularly important reason for training models on CHILDES is that, in prior work, different types of training data have yielded diverging results: Recent  models trained on synthetic data failed to properly acquire yes/no questions \cite{mccoy2020trees,petty2021transformers}, whereas ones trained on large Internet corpora scored well on evaluations of yes/no questions \cite{lin2019open,warstadt2020neural}. 
    Given these differing results, it is not clear from past work how these models would generalize when faced with the type of data that children receive.

\section{Overview of Experimental Setup}

We evaluated models on yes/no questions in two ways. 
First, we used relative acceptability judgments (Experiment 1): We trained neural networks on the task of language modeling (predicting the next word at every point in the sentence) and  evaluated whether they assigned a higher probability to sentences consistent with \textsc{LinearQ} or \textsc{HierarchicalQ}. 
Our second approach was based on text generation (Experiment 2): We trained networks to take in a declarative sentence and output the corresponding question, and tested whether they generalized in a way more consistent with  \textsc{LinearQ} or \textsc{HierarchicalQ}. 
Under both framings, we trained models on data from CHILDES and evaluated them on targeted datasets constructed to differentiate \textsc{LinearQ} and \textsc{HierarchicalQ}.

\section{Experiment 1: Relative Acceptability}\label{sec:experiment-1}
 
    \subsection{Dataset} 
    To train models on data as similar as possible to the sentences children receive, we extracted data from CHILDES  \cite{macwhinney2000childes}. 
    We used the North American English portion.
    We wished to replicate children's \textit{input}, so we excluded the children's own utterances, leaving a 9.6-million-word corpus. 
    We allocated 90\% of the data to training, 5\% to validation, and 5\% to testing. 
    We replaced words that appeared two or fewer times in the training set with $<$unk$>$, giving a replacement rate of 0.3\%. 
    See Appendix \ref{app:preprocessing} for more details.

    \subsection{Task: Next-Word Prediction}\label{scc:experiment_1_task}
    
    We trained models on next-word prediction, also known as language modeling. 
    We chose this task for two reasons.
    First, it is clear empirically that next-word prediction can teach neural networks a substantial amount about syntax \cite[e.g.,][]{hu2020systematic}. 
    Second, it is plausible that humans perform some version of next-word prediction during sentence processing \cite{altmann1999incremental,hale2001probabilistic,levy2008expectation,kutas2011look} and that such prediction may play a role in acquisition \cite{elman1991distributed}.
    Thus, while next-word prediction is certainly not the only goal of human language learners, we view this task as a reasonable first step in emulating human language acquisition.
    
    \subsection{Architectures}
    We used two neural network architectures: LSTMs \cite{hochreiter1997long} and Transformers \cite{vaswani2017attention}. 
    We chose these models for two reasons. First, they have been the most successful architectures in NLP. Thus, we have reason to believe that, of the types of low-bias models invented, these two are the ones most likely to discover linguistic regularities in our CHILDES training data.
    Second, the two architectures process sequences very differently (via recurrence vs.\ via attention). 
    Thus, if both generalize similarly, we would have evidence that what was learned is strongly evidenced in the data, rather than due to a quirk of one particular architecture. 
        
    For our LSTMs, we used 2 layers, a hidden and embedding size of 800, a batch size of 20, a dropout rate of 0.4, and a learning rate of 10.
    For our Transformers, the corresponding values were 4, 800, 10, 0.2, and 5, and we used 4 attention heads. 
    We chose these values based on a hyperparameter search described in Appendix \ref{app:hyperparameter_search}. 
    All following results are averaged across 10 runs with different random seeds.

     \subsection{Results: Language Model Quality}       
        Before testing models on  questions, we used perplexity to evaluate how well they captured the basic structure of their training domain. 
        As a baseline, we used a 5-gram model with Kneser-Ney smoothing \cite{kneser1995improved} trained with KenLM \cite{heafield2011kenlm}. 
        The test set perplexity for the 5-gram baseline was 24.37, while the average test set perplexity for the LSTMs and Transformers was 20.05 and 19.69, respectively.
        For perplexity, lower is better. Thus, both neural network types outperformed the strong baseline of a smoothed 5-gram model, showing that they performed well at capturing the basic statistics of their training domain.\footnote{For an intuitive illustration of our model quality, see the sample text generated by them in Appendix~\ref{app:generated_text}.}

    \subsection{General Syntactic Evaluation}\label{sec:zorro}
        
        As an additional way to check the validity of our setup, we evaluated our models on the Zorro dataset  \cite{huebner2021babyberta}, which is based on BLiMP \cite{warstadt2020blimp}. Zorro contains 24 evaluations, each of which targets one syntactic phenomenon (e.g., subject-verb agreement) and involves sentence pairs for which one sentence is grammatical, and the other is minimally different but ungrammatical (e.g., by violating subject verb agreement). 
        A model is said to get a sentence pair correct if it assigns a higher probability to the grammatical sentence than the ungrammatical one.
        \citet{huebner2021babyberta} showed that Transformers trained on CHILDES data can perform well on many of the Zorro categories, so if our setup is sound, our own models should also perform well on Zorro.

        See Appendix \ref{app:babyberta} for full results. For each syntactic phenomenon, most model re-runs scored above 0.9, though at least one scored near the chance level of 0.5. For each re-run of each architecture there is at least one phenomenon for which the model scores over 0.97, and many models score 1.00 on some phenomena. 
        Thus, all models score well on at least some syntactic evaluations, attaining results comparable to those of \citet{huebner2021babyberta} and providing additional support for the validity of our setup. We now test whether these models have also successfully learned the specific phenomenon that we focus on, yes/no questions---a phenomenon not included in the Zorro dataset.

     \subsection{Yes/No Questions}\label{sec:yn_questions_results}

     \paragraph{Evaluation Dataset: Forced-Choice Acceptability Judgments}

     As a first way to test whether our models have learned \textsc{HierarchicalQ}, we evaluate whether they assign higher probabilities to sentences consistent with \textsc{HierarchicalQ} than to minimally different sentences that are ungrammatical.
     For this purpose, we create an evaluation dataset
     containing groups of 6 questions, each created by starting with a declarative sentence, such as \ref{ex:6fc_origin}, and then 
     deleting the \color{red}\textbf{first}\color{black}, \color{blue}\textbf{\underline{main}}\color{black}, or neither auxiliary, and inserting the \color{red}\textbf{first} \color{black} or \color{blue}\textbf{\underline{main}} \color{black} auxiliary at the front of the sentence.\footnote{It would be possible to also use a `prepose other' category, where an auxiliary not in the input is inserted \cite{mccoy2018revisiting}. We excluded this category because using it would raise complications about which `other' auxiliary to choose.} 
     For instance, in \ref{ex:6fc_p1d2}, the \color{red}\textbf{first} \color{black} auxiliary has been preposed, and the \color{blue}\textbf{\underline{main}} \color{black} auxiliary has been deleted. 
     
     \ex. The dog who \color{red}\textbf{has} \color{black} seen a boy \color{blue}\textbf{\underline{did}} \color{black} try.\label{ex:6fc_origin}
     
     \vspace{-6pt}
     \ex. \label{ex:6fc}
     \a.    \color{red}\textbf{Has} \color{black} the dog who seen a boy \color{blue}\textbf{\underline{did}} \color{black} try?\label{ex:6fc_p1d1}
     \b.    \color{red}\textbf{Has} \color{black} the dog who \color{red}\textbf{has} \color{black} seen a boy try?\label{ex:6fc_p1d2}
     \c.    \color{red}\textbf{Has} \color{black} the dog who \color{red}\textbf{has} \color{black} seen a boy \color{blue}\textbf{\underline{did}} \color{black} try ?\label{ex:6fc_p1d0}
     \d.    \color{blue}\textbf{\underline{Did}} \color{black} the dog who seen a boy \color{blue}\textbf{\underline{did}} \color{black} try?\label{ex:6fc_p2d1}
     \e.    \color{blue}\textbf{\underline{Did}} \color{black} the dog who \color{red}\textbf{has} \color{black} seen a boy try?\label{ex:6fc_p2d2}
     \f.    \color{blue}\textbf{\underline{Did}} \color{black} the dog who \color{red}\textbf{has} \color{black} seen a boy \color{blue}\textbf{\underline{did}} \color{black} try?\label{ex:6fc_p2d0}
     
     Within each group, we evaluate which question the model assigned the highest probability to. If a model has correctly learned \textsc{HierarchicalQ}, it should assign the highest probability to the question consistent with this rule, such as \ref{ex:6fc_p2d2}.
     
     Several past papers about yes/no questions have used the same general approach  \cite{lewis2001learnability,reali2005uncovering}. However, these papers considered only pairs of sentences, whereas we consider groups of 6 to allow for a wider range of possible generalizations that a model might have learned.

     To generate the declaratives from which we formed groups of 6 questions, we used the context-free grammar (CFG) in Appendix \ref{app:cfgs}, which has a vocabulary selected from the most common words in CHILDES.
     Each declarative generated by the CFG (e.g., \ref{ex:6fc_origin}) contains two auxiliary verbs: one before the sentence's main verb and one inside a relative clause modifying the subject. 
     One potential problem is that some questions are consistent with both \textsc{HierarchicalQ} and \textsc{LinearQ}. 
     For instance, \ref{ex:ambig_q} can be formed from \ref{ex:ambig_correct} with the \textsc{HierarchicalQ}-consistent steps \textsc{Prepose-Main,Delete-Main}, or from \ref{ex:ambig_wrong} with the  \textsc{LinearQ}-consistent steps \textsc{Prepose-First,Delete-Main}. 
     
      \ex. \a. Did the boy who did see the person laugh?\label{ex:ambig_q}
     \b. The boy who did see the person did laugh.\label{ex:ambig_correct}
     \c. The boy who did see the person can laugh.\label{ex:ambig_wrong}
     
     To avoid this problem, we required that the auxiliary before the main verb must select for a different verb inflection than the one in the relative clause.  For instance in \ref{ex:6fc_origin}, \textcolor{blue}{\textbf{did}} selects for the verb's bare form, while \textcolor{red}{\textbf{has}} selects for the past participle form.
     Thus, the auxiliary at the start of the question could only correspond to whichever auxiliary in the declarative has the same selectional properties.\footnote{A model could succeed on this dataset with a rule that relates the auxiliary at the start of a question with the \textit{last} auxiliary in the declarative form. Since our models fail on this dataset, this consideration is not relevant here.}

     \paragraph{Results: Relative Question Acceptability}
    
\begin{figure}[t!]
        \begin{centering}
        \includegraphics[width=\columnwidth]{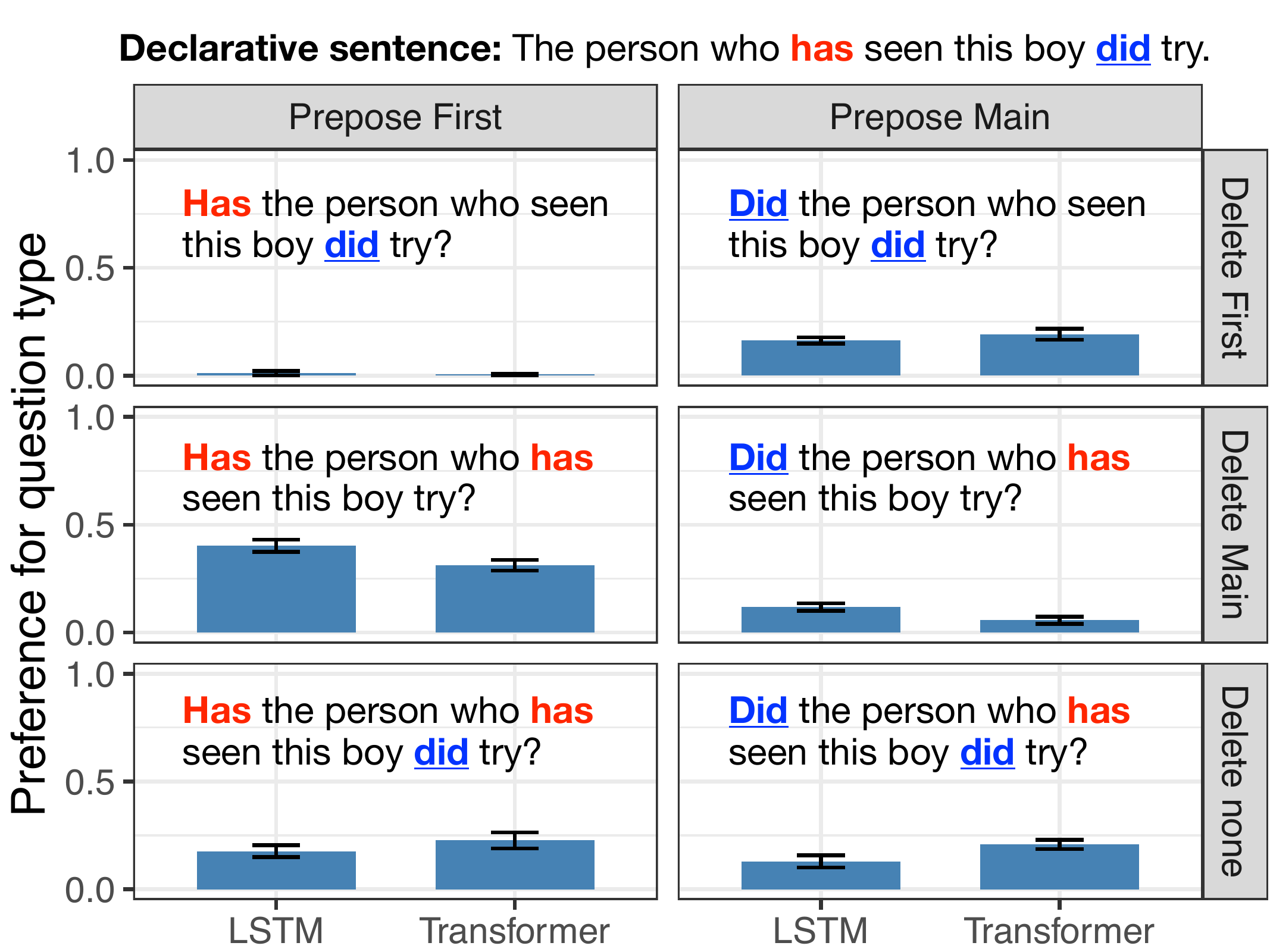}
        \caption{The question types that models prefer when offered a choice between 6 questions.
        These 6 questions are formed by modifying a declarative with a relative clause on the subject according to `prepose' and `delete' rules. The correct category is \textsc{Prepose Main, Delete Main}. Within each architecture, the proportions across all 6 question types necessarily sum to 1. Each bar shows the average across 10 model re-runs, with single-standard-deviation error bars.}
        \label{fig:prepose_delete} 
        \end{centering}
    \end{figure}

     For each sentence group, we used per-word perplexity to see which of the 6 candidates the models scored most highly.\footnote{We also explored evaluation of the models with a more complex measure called SLOR where we additionally normalized scores by word frequency \cite{pauls2012large}. Both metrics produced qualitatively similar results, so we only report the simpler metric here. See Appendix \ref{app:slor}.}  For both LSTMs and Transformers, the correct category (\textsc{Prepose Main, Delete Main}) was the second-rarest choice,
     and the most frequent preference was for \textsc{Prepose First, Delete Main}, a category that is only partially correct because it references linear order in addition to hierarchical structure  (Figure~\ref{fig:prepose_delete}).
     
     Thus, neither model displays preferences consistent with the correct, fully-hierarchical generalization.
     The two model types showed similar scores, which may mean that these results are largely driven by the statistics of the training data that both models share, rather than the models' differing inductive biases.

     One of the incorrect categories---\textsc{Prepose Main, Delete None}, such as \ref{ex:6fc_p2d0}---only requires reference to hierarchical structure, so it could be said to capture the hierarchical nature of yes/no questions.
     Nonetheless, this category was also relatively rare: combining the two fully hierarchical possibilities (\textsc{Prepose Main, Delete Main} and \textsc{Prepose Main, Delete None}) accounts for only 26\% of LSTM preferences and 27\% of Transformer preferences, meaning that both models over 70\% of the time favored a sentence generated at least partially based on linear order.
     
     There are two likely reasons for why our models performed so poorly on yes-no questions when they performed well on many of the phenomena in the Zorro dataset (Section \ref{sec:zorro}). First, yes/no questions may simply be harder to learn than the other phenomena; indeed, yes/no questions are often singled out as being likely to pose difficulties for a general-purpose learner (Section \ref{sec:intro}). While this focus in prior literature might simply be a historical coincidence, it is also possible that it points to a true difference in ease of learning. Alternatively, it might be that the six-way evaluation we used for yes/no questions is stricter than the binary judgments used for the Zorro dataset.

\section{Experiment 2: Question Formation}

    The previous experiment was designed to operate entirely in the next-word-prediction paradigm, motivated by arguments from past literature about the strength and relative ecological validity of next-word-prediction as a training objective (see Section~\ref{scc:experiment_1_task}). However, one of this setup's shortcomings is that \textsc{HierarchicalQ} describes \underline{correspondences} between questions and declaratives, but Experiment 1 focused on questions \underline{alone}, with no consideration of declaratives.
    
    In this second experiment, to better capture that \textsc{HierarchicalQ} is defined over sentence pairs, we trained models on a sentence-pair task: transforming a declarative into a question \cite{mccoy2020trees}.
    For instance, given \textit{the child did learn} the model must produce \textit{did the child learn ?}

    We evaluated models in two ways.
    First, we checked if the models' predictions fully matched the correct questions. This full-sentence evaluation is demanding, and models might 
    fail this evaluation
    for reasons unrelated to our core hypotheses. For instance, given \textit{the child did learn} the model might produce \textit{did the baby learn}, which would be marked as incorrect, even though this lexical error is not relevant to \textsc{HierarchicalQ}.
    
    As a metric that is less demanding and that also more directly targets \textsc{HierarchicalQ}, we measured if the first word of the output question corresponded to the first or main auxiliary of the input. Critically, \textsc{LinearQ} and \textsc{HierarchicalQ} make different predictions for the first word of a question so long as the two auxiliaries are distinct: see \ref{different_predictions}. 
    Because this framing lets the model freely generate its output (instead of choosing one option from a pre-specified set), we allow for the possibility that the rule learned by models may not be identical to any of our manually-generated hypotheses. 

    Solely training models to perform this transformation involves the implicit assumption that, when children acquire English yes/no questions, the only evidence they leverage is English yes/no questions. However, other types of sentences may also provide useful evidence \cite{pearl2016indirect}: e.g., \textit{wh}-questions also illustrate subject-auxiliary inversion \cite{pullum2002empirical}, while, more generally, many types of sentences could provide evidence that syntax as a whole is hierarchical  \cite{perfors2011learnability}. To explore this possibility,
    we compared a condition in which models were only trained to perform question formation (the \textsc{Question Formation} condition) to another in which models were first pre-trained on next-word prediction with the exact same setup as in Experiment~1 before being further trained to perform question formation (the \textsc{Next-word Prediction + Question Formation} condition).

 \subsection{Dataset}
    \paragraph{Training Set} Our question formation dataset consisted of the yes/no questions in the CHILDES Treebank \cite{pearl2013computational,pearl2013syntactic}, a parsed subset of CHILDES containing 189,359 sentences. 
    We used these parses to extract all yes/no questions from the CHILDES Treebank and derive their corresponding declarative forms.
    The resulting declarative was concatenated with the question. 
    An example declarative/question pair is:
    
        \ex. \small \texttt{you can spell your name . can you spell your name ?} \label{ex:decl-quest-pair}
    
    \noindent
    The training set consisted of 10,870 declarative/question pairs, the validation set 1,360 pairs, and the test set 1,358 pairs (we will call this test set the \textit{randomly-partitioned test set} to distinguish it from two other evaluation sets discussed below). 
    We trained models to perform next-word prediction on such concatenated sentence pairs. 

    The first-word accuracy of the trained model was then computed based on the model's prediction for the word after the period in each test example, while the full-sentence accuracy was computed based on its predictions for all tokens after the period.
    All questions in the randomly-partitioned test set were withheld from both the question-formation training set and the next-word-prediction training set. Thus, models had not seen these test examples in their training, even in the \textsc{Next-word Prediction + Question Formation} condition in which they were trained on both tasks.

    \paragraph{Evaluation Sets} In addition to the randomly-partitioned test set, we used CFGs to generate two targeted evaluation sets.
    As in Experiment 1, we selected the CFGs' vocabulary from common words in our CHILDES data. 
    In sentences generated from the first CFG, the sentence's first auxiliary was also its main auxiliary, so  \textsc{LinearQ} and \textsc{HierarchicalQ} make the same predictions. \ref{ex:first-equal-to-main} exemplifies the type of declarative-question pair in this dataset. We call this dataset \textsc{First-Aux} = \textsc{Main-Aux}.
    For sentences generated by the second CFG, the main auxiliary was the \textit{second} auxiliary in the sentence; thus, these examples disambiguate \textsc{LinearQ} and \textsc{HierarchicalQ}. Example \ref{ex:first-not-equal-to-main} is a declarative-question pair from this evaluation set. We call this dataset \textsc{First-Aux} $\not =$ \textsc{Main-Aux}. 
    See Appendix \ref{app:cfgs} for the CFGs used. 

    \setlength{\Exlabelwidth}{0.4em}
    \ex. \a. \small \texttt{a girl was playing . was a girl playing ?}\label{ex:first-equal-to-main}
    \b. \small \texttt{a boy who is playing can try . can a boy who is playing try ?} \label{ex:first-not-equal-to-main}
    
    \setlength{\Exlabelwidth}{0.7em}
    
    %We sampled 10,000 declarative sentences from these grammars and transformed them into questions according to \textsc{HierarchicalQ} to create our evaluation sets. 
    
    \subsection{Results}\label{sec:question_formation_results}
        \paragraph{Randomly-Partitioned Test Set} 
        
        The LSTMs and Transformers in the \textsc{Question Formation} condition performed well on the randomly-partitioned test set, with a full-question accuracy of 0.68 $\pm$ 0.014 and 0.87 $\pm$ 0.005 (averaged across 10 reruns with margins indicating one standard deviation). 
        The models in the \textsc{Next-word Prediction + Question Formation} condition performed similarly well, with a full-question accuracy of 0.66 $\pm$ 0.008 for the LSTMs and 0.93 $\pm$ 0.004 for the Transformers. 
        For both model types, the  first-word accuracy for the question was nearly 1.00 across re-runs.
        We suspect that Transformers have a  stronger full-question accuracy because producing the question requires copying all words from the declarative (but in a different order).
        Copying is likely easy for Transformers because they can attend to specific words in the prior context, while our LSTMs must compress the entire context into a fixed-size vector, which may degrade the individual word representations.
        Because both model types achieved near-perfect performance on the crucial first-word accuracy metric, we conclude that our models have successfully learned how to handle the types of declarative/question pairs that we extracted from the CHILDES Treebank. 
        
\begin{figure}[t!]
        \begin{centering}
        \includegraphics[width=\columnwidth]{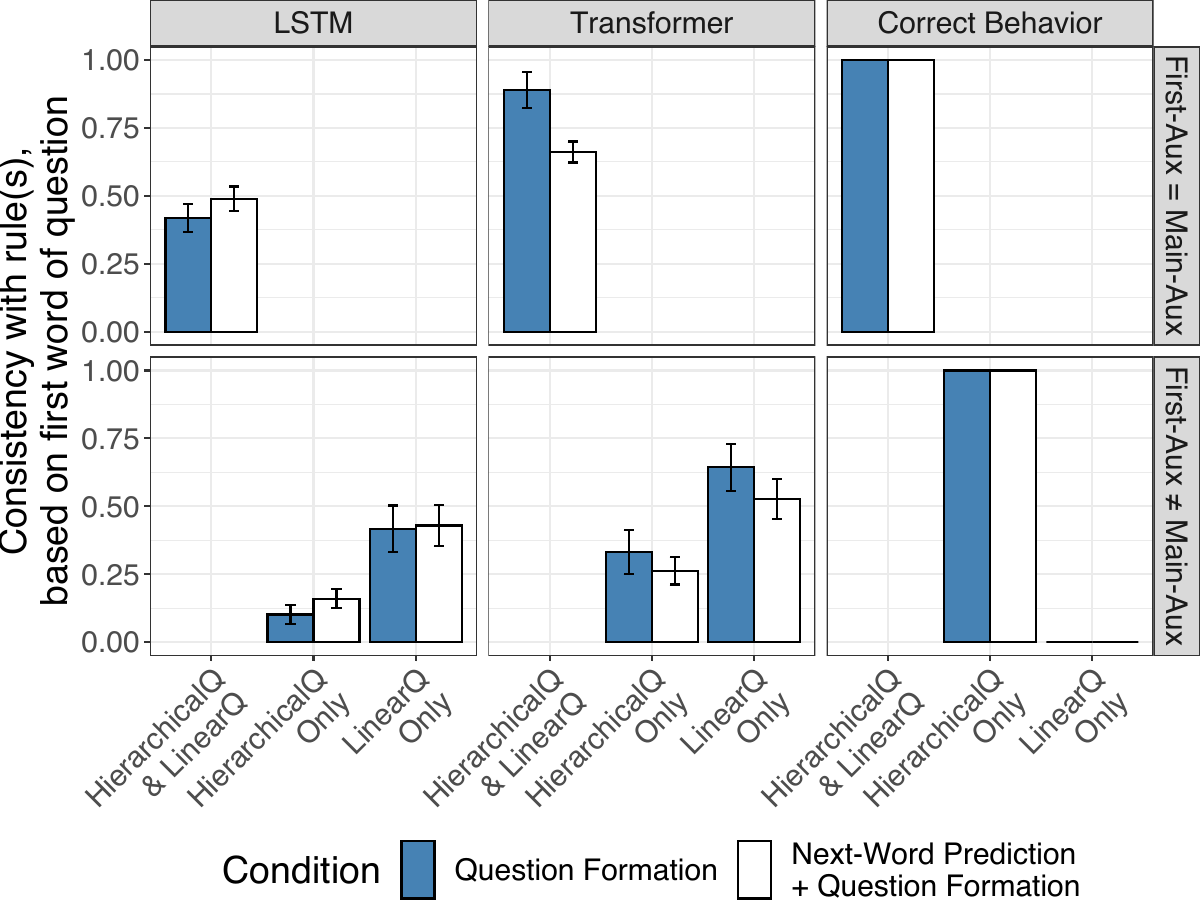}
        \caption{Proportion of model-produced or target questions that were consistent with the linear rule \textsc{LinearQ} and/or the hierarchical rule \textsc{HierarchicalQ}.  In the \textsc{First-Aux = Main-Aux} dataset, the first auxiliary is the main auxiliary, so both \textsc{LinearQ} and \textsc{HierarchicalQ} produce the correct question string. The \textsc{First-Aux $\neq$ Main-Aux} dataset disambiguates the two rules. Each bar in the LSTM and Transformer facets shows the average across 10 model re-runs, with error bars showing one standard deviation. 
        The Correct Behavior column shows how a model would perform if it had perfectly learned the correct generalization.
        }
        \label{fig:results}
        \end{centering}
        \end{figure}

        \paragraph{Targeted Evaluation Sets}
        On our targeted evaluation sets, models seldom produced the complete question correctly. 
        On the more lenient measure of first-word accuracy, for cases where \textsc{LinearQ} and \textsc{HierarchicalQ} predict the same first output word (\textsc{First-Aux = Main-Aux}), 
        the Transformer trained only on question formation performed strongly, while the Transformer trained on both tasks, and both LSTMs, performed decently (Figure \ref{fig:results}; note chance performance is 1/vocabulary size, which is near 0.00).
        For cases that disambiguate the two rules (\textsc{First-Aux $\neq$ Main-Aux}), both models in both conditions performed more consistently with \textsc{LinearQ} than \textsc{HierarchicalQ}.
        Training on next-word prediction before question formation had inconsistent effects: it modestly increased the chance of hierarchical behavior in LSTMs, and decreased it in Transformers.

\begin{figure}[t!]
        \begin{centering}
        \includegraphics[width=\columnwidth]{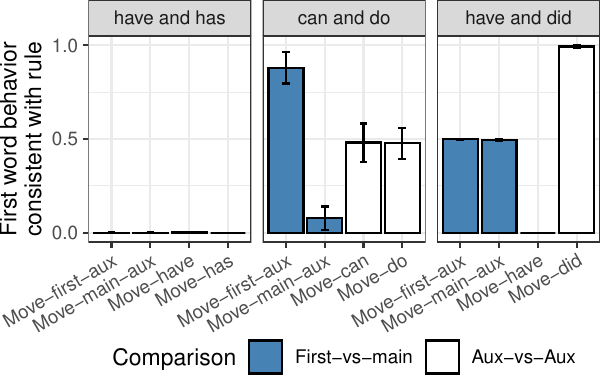}
         \caption{Lexical specificity in model behavior. Each facet considers only the evaluation examples containing the two auxiliaries in the facet heading; e.g., the \textit{can and do} facet includes, for example, the inputs \textit{the children who \textbf{can} play \textbf{do} learn} and \textit{the children who \textbf{do} play \textbf{can} learn}. 
         The bars show the proportion of model predictions for the first word of the output that are consistent with four potential movement rules, averaged across 10 model re-runs and with error bars showing one standard deviation above and below the mean. This plot only shows an illustrative subset of auxiliary pairs for one model type (Transformers in the \textsc{Next-Word Prediction + Question Formation} condition); see Appendix \ref{app:lex_id} for the full results.}
         
         \label{lexically-specific-key-comparisons}
        \end{centering}
        \end{figure}

        \paragraph{Lexical Specificity} In Appendix \ref{app:lex_id}, we further break down the \textsc{First-Aux $\neq$ Main-Aux} results based the auxiliaries' identity. The generalization pattern varied considerably across auxiliary pairs.  For some auxiliary pairs, the auxiliary chosen to begin the question was usually neither auxiliary in the input (Figure \ref{lexically-specific-key-comparisons}, left facet). 
        For other pairs, models usually chose the first auxiliary, regardless of lexical identity (Figure \ref{lexically-specific-key-comparisons}, middle facet).
        Finally, for some pairs, the auxiliary chosen was usually the same one, regardless of whether it was the first or main auxiliary (Figure \ref{lexically-specific-key-comparisons}, right facet).

        Generalization based on lexical identity is rarely  considered in past discussions of English yes/no question acquisition. Of the papers on this phenomenon (see \citet{clark2010linguistic}, \citet{lasnik2017argument}, and \citet{pearl2021poverty} for overviews), the only one to our knowledge that discusses lexical specificity is \citet{frank2007transformational}, which studied models trained on synthetic data. Our results highlight the importance of testing for a broad range of generalizations:  Lexically-specific hypotheses appear attractive for our low-bias learners, so an account of what biases can yield human-like learning should rule out these lexically-specific hypotheses along with linear ones.
        
\section{Discussion}

We have found that, when trained on child-directed speech, two types of standard neural networks performed reasonably well at capturing the statistical properties of the dataset, yet their handling of English yes/no questions was more consistent with a linear rule \textsc{LinearQ} than the correct hierarchical rule \textsc{HierarchicalQ}. These results support the hypothesis that a learner requires a hierarchical bias to consistently learn hierarchical rules when learning from the linguistic data children receive. 

\subsection{Takeaways for LSTMs and Transformers}

When trained on massive corpora, LSTMs and Transformers perform impressively on some syntactic evaluations.
Based on such results, it is tempting to conclude that the general-purpose biases of these architectures suffice to yield human-like syntax acquisition.
Our results caution against this interpretation: When we trained the same architectures on data more similar to children's input, they failed to learn the structure of English yes/no questions.
Thus, at least when learning from text alone, LSTMs and Transformers do not display human-like language learning---they do not generalize as humans do \textit{from the data that humans receive}.

\subsection{Takeaways for the Poverty of the Stimulus Debate}

Below we specify four possible  positions in the poverty-of-the-stimulus debate about the adequacy of children's input for inducing hierarchical rules in low-bias learners, arranged from assuming the most limited to the most expansive innate component:

\ex. \textbf{Any inductive biases:} Any learner trained on CHILDES will generalize like humans do.\label{ex:position1}

\ex. \textbf{Any inductive biases that enable in-distribution learning:} Any learner that captures the statistical patterns of the training distribution will generalize to \textsc{HierarchicalQ}.\label{ex:position2}

\ex. \textbf{Some non-hierarchical inductive biases:} Some general-purpose learners will generalize as humans do, but others will not.\label{ex:position3}

\ex. \textbf{Only a hierarchical inductive bias:} No general-purpose learners will generalize as humans do:  hierarchical biases are necessary.\label{ex:position4}

Position \ref{ex:position1} is clearly false: many learners cannot learn certain aspects of syntax, no matter their training data (e.g., bigram models cannot capture long-distance dependencies).
Our work shows that position \ref{ex:position2} is also false: Though our models performed well on the in-distribution test sets of Experiments 1 and 2, they did not generalize in human-like ways. This leaves positions \ref{ex:position3} and \ref{ex:position4}, which our existing results cannot differentiate. It is possible that only learners with hierarchical inductive biases can demonstrate human-like language learning (position \ref{ex:position4}), but also  that some learners without this bias can succeed (position \ref{ex:position3})---just not the learners we tested. 
For further discussion of how computational modeling can bear on learnability arguments, see \citet{wilcox2021using}.

One potential solution supporting position \ref{ex:position3} would be that learners leverage the hierarchical structure of some syntactic phenomenon to help conclude that other, impoverished phenomena are hierarchical \cite{perfors2011learnability,mulligan2021structure}. However, our results from Experiment 2 show that giving learners access to a wider range of phenomena does not automatically improve hierarchical generalization: Models' performance on question formation was not substantially improved (and in some cases was even harmed) when they were trained not just on question formation but also on next-word prediction on the entire CHILDES corpus. Thus, although training on text that contains many linguistic phenomena 
can give models a hierarchical inductive bias when the training is done over large Internet corpora \cite{warstadt2020neural,mueller2022coloring}, our results provide evidence that this conclusion does not extend to models trained on child-directed speech.

Though both \ref{ex:position3} and \ref{ex:position4}
remain as possibilities, we believe that our results more strongly support \ref{ex:position4}.
Of all currently available general-purpose learners, LSTMs and Transformers are the best at modeling the probabilistic structure of linguistic data. Therefore, if child-directed speech contains clear evidence for the hierarchical nature of yes/no questions---evidence so clear that at least some general-purpose learners could recognize it---it is likely that LSTMs and Transformers would be among the set of general-purpose learners that could use this evidence to make hierarchical generalizations in our experiments. The fact that these architectures instead predominantly favored linear generalizations therefore supports position \ref{ex:position4}.

\subsection{How to test for \textsc{HierarchicalQ}}

We have argued that an ideal simulation of the acquisition of English yes/no questions would have the following properties:

\ex. The training data should be similar to children's linguistic input.\label{ex:property1}

\ex. The training task should be ecologically valid.\label{ex:property2}

\ex. The evaluation method should focus on correspondences between pairs of sentences rather than the acceptability of individual sentences.\label{ex:property3}

Property \ref{ex:property1} motivated our use of text from CHILDES as the training data.
We are not aware of a single experimental setup that fully satisfies both Property \ref{ex:property2} and Property \ref{ex:property3}, so we instead used two experiments, each one focusing on one property at the cost of satisfying the other one less well. Experiment 1 works entirely in the context of the relatively ecologically valid task of next-word prediction, motivated by Property \ref{ex:property2}, but its evaluation is only based on the acceptability of individual sentences, failing to satisfy Property \ref{ex:property3}. Experiment 2 fully satisfies Property \ref{ex:property3} by using an evaluation based on sentence pairs, at the cost of including a less ecologically-valid training component based on sentence transformations. Both experiments yielded qualitatively similar conclusions (failure of models to learn \textsc{HierarchicalQ}).

\subsection{Quantity of Training Data}
    
The size of our training set was plausibly in the range from which children can acquire \textsc{HierarchicalQ}.
\citet{crain1987structure} found that children between ages 3 and 5 behaved much more consistently with \textsc{HierarchicalQ} than \textsc{LinearQ}.
Though these children 
made many errors,
their errors were usually compatible with a hierarchical rule (e.g., \textsc{Prepose Main, Delete None} errors: see Section \ref{sec:yn_questions_results}).
By age 3, American children receive approximately 10 to 33 million words of input \cite{hart1995meaningful}, and the 8.5 million of our training set is near the lower end of that range. Thus, while we cannot be completely certain, it is reasonable to suppose  that a learner that generalizes as children do would favor \textsc{HierarchicalQ} after being trained on our training set. Our models, in contrast, preferred sentences generated in ways based on linear order (Figures \ref{fig:prepose_delete} and \ref{fig:results}), a error category very rare in children  \cite{crain1987structure,ambridge2008structure}.

In order to give our models the strongest chance of generalizing correctly, it would have been ideal to provide a quantity of data closer to 33 million words, the high end of \citeauthor{hart1995meaningful}'s range. 
Our data source did not contain enough text to make this possible, but future work could investigate ways to augment the data using other sources.

\subsection{Type of Training Data}\label{sec:input}

Our training set was both qualitatively and quantitatively closer to children's input than the massive Internet corpora standardly used to train models in NLP \cite{linzen2020accelerate}. This difference is important: \citet{lin2019open}, \citet{warstadt2020neural}, and \citet{mueller2022coloring} all found evidence that models trained on large Internet corpora performed well on yes/no questions evaluations, whereas our models trained on CHILDES performed poorly---though we cannot be certain the differences in results are solely due to differences in the training data, since these prior papers  used different model architectures, training tasks, and evaluation setups. 

Though our training data are more similar to children's input than massive Internet corpora are, differences remain. 
Our experiments omit several aspects of a child's experience that might help them acquire syntax, such as prosody \cite{morgan1996signal}, visual information \cite{shi2019visually}, meaning \cite{fitz2017meaningful,abend2017bootstrapping}, and social interaction \cite{kuhl2003foreign,rowe2020context}, all of which involve information that might correlate with syntactic structure and thus provide cues to the correct hierarchical generalization. On the other hand, our dataset might present an easier learning scenario than children are faced with, because children must learn to segment the speech stream into words \cite{lakhotia2021generative}, while our models do not need to. 
Further, 
though real-world grounding could provide 
helpful
information, learners might struggle to leverage this information 
due to difficulty
determining what is being discussed in the physical world \cite{gleitman2005hard}.

\section{Conclusion}

In this work, we trained two types of neural networks (LSTMs and Transformers) on sentences of the types available to children and then analyzed what they had learned about English yes/no questions.
Across several evaluation paradigms, these models failed to generalize in human-like ways: Humans display hierarchical generalization, while the models' generalization was
 instead based on linear order and individual words' identities.  Our results support the hypothesis that human-like linguistic generalization requires biases stronger than those of LSTMs and Transformers.
Future work should investigate what inductive biases enable successful generalization. One approach would be to test architectures with built-in hierarchical structure; past work has shown that such architectures have a hierarchical bias \cite{mccoy2020trees} and generalize better on the hierarchical phenomenon of subject-verb agreement \cite{kuncoro2018lstms,lepori2020representations}, so they may also generalize better on English yes/no questions. 
A final direction 
would be to expand the input beyond words alone so that learners can leverage hierarchical structure that is present in 
other modalities, such as hierarchical structure in visual scenes.

\section*{Ethics Statement}

\paragraph{Use of human data:} While we did not collect any new human data ourselves, many of our analyses involved the use of prior datasets within the CHILDES database. All of these datasets were collected in accordance with IRB policies at the institutions of the data collectors, and all followed standard practices in obtaining informed consent and deidentifying data.\footnote{\url{https://talkbank.org/share/irb/}}

\section*{Limitations} 
We view strong performance on our evaluation datasets as necessary but not sufficient to demonstrate human-like learning. Thus, if models perform poorly on our datasets (as the models we evaluated did), then we have strong reason to conclude that models are not learning in human-like ways. If future models perform better, such results would be consistent with human-like learning but would not conclusively establish that models learn as humans do, as they might instead be using some shallow heuristic that is not controlled for in our datasets. In other words, a criterion that is necessary but not sufficient facilitates strong conclusions about failure but does not facilitate strong conclusions about success. If future papers are faced with models that are more successful, such papers would ideally supplement results based on our datasets with analyses of models' internal strategies in order to more conclusively establish that what they have learned is not a spurious heuristic.

Thus an important risk of our proposed analyses is that future work using the same analyses might draw overly strong conclusions based on increased model performance, leading to overestimates of model strength. Such overestimates are an issue because they can lead users to place more trust in a model than is warranted.

\section*{Acknowledgments}

For helpful comments and discussion, we are grateful to Najoung Kim, An Nguyen, Grusha Prasad, Paul Smolensky, Paul Soulos, and the NYU Computation and Psycholinguistics Lab. 
Any errors are our own.
We are also grateful to the Maryland Advanced Research Computing Center (MARCC) for providing the computing resources used in our experiments.

Portions of this research were supported by the National Science Foundation (NSF) under grants BCS-2114505, BCS-1919321, and Graduate Research Fellowship Program grant no.\ 1746891. Any opinions, findings, and conclusions or recommendations expressed in this material are those of the authors and do not necessarily reflect the views of the National Science Foundation.

\bibliography{anthology,custom}
\bibliographystyle{acl_natbib}

\appendix

    \section{CHILDES preprocessing details}\label{app:preprocessing}
    
    The train, test, and validation split kept each document in the corpora intact to allow for learning of context. 
    Since a document roughly correspond to a single recording session, and the sentence order within each document was not randomized, the networks could utilize cross sentence context while predicting the next word.
    
    Generally, we kept the data as close to the actual input that the child receives as possible. 
    However, in some cases we modified tokenization to match the CHILDES Treebank, a syntactically parsed subset of the CHILDES corpora. 
    For instance, contractions were split, e.g. we replaced \textit{don't} with \textit{do n't},

    The ages of the children  vary by corpus, ranging from six months to twelve years. Almost 95\% (49/52) of the corpora consist of transcriptions with children between one and six years of age. 
    
    Note that for Experiment 2, we used the same  vocabulary as we used in Experiment 1, which means that the words that were not present in Experiment 1's vocabulary were replaced with $<$unk$>$ tokens.

    The unprocessed CHILDES datasets were downloaded in XML format from the online XML version\footnote{\url{https://childes.talkbank.org/data-xml/}} of the CHILDES database \cite{macwhinney2000childes}.\footnote{\url{https://childes.talkbank.org}} A modified NLTK CHILDESCorpusReader\footnote{\url{https://www.nltk.org/howto/childes.html}} was used to parse the XML into plain text for training. 
    
    The CHILDES dataset is licensed for use under a CC BY-NC-SA 3.0 license\footnote{\url{https://talkbank.org/share/rules.html}}. Under the terms of this license, the data can be freely used and adapted, as long as it is not used for commercial purposes and as long as attribution is provided.\footnote{\url{https://creativecommons.org/licenses/by-nc-sa/3.0/}} Our usage fits these criteria.
    
    Though CHILDES contains many corpora of many languages, we use only corpora from the North American English subset of CHILDES, which contains child-directed speech with many different North American children. See the CHILDES database for more details. 
    
    By the CHILDES rules for data citation,\footnote{\url{https://talkbank.org/share/citation.html}} research that relies on more than 6 of the corpora need only cite the overall database, not each individual corpus. 
    
    All the data on CHILDES must adhere to IRB guidelines,\footnote{\url{https://talkbank.org/share/irb/}} including a requirement for anonymity.
    
    The final dataset is included in our GitHub repository. This dataset is not intended for commercial use.
    
    \paragraph{CHILDES corpora included} The CHILDES corpora that we used were: Bates, Bernstein, Bliss, Bloom70, Bloom73, Bohannon, Braunwald, Brent, Brown, Carterette, Clark, Cornell, Demetras1, Demetras2, EllisWeismer, Evans, Feldman, Garvey, Gathercole, Gelman, Gillam, Gleason, HSLLD, Haggerty, Hall, Higginson, Kuczaj, MacWhinney, McCune, McMillan, Morisset, NH, Nelson, NewEngland, NewmanRatner, Normal, POLER, Peters, Post, Rollins, Sachs, Sawyer, Snow, Soderstrom, Sprott, Suppes, Tardif, Valian, VanHouten, VanKleeck, Warren, Weist. 
    
\section{Hyperparameter Search and Model Implementation}\label{app:hyperparameter_search}

    We conducted a hyperparameter search for each of the architectures we investigated (LSTMs and Transformers). Our broad goal in this paper is to investigate the extent to which capturing the statistical properties of the CHILDES dataset naturally leads a learner to capture the structure of yes/no questions. Therefore, we sought to find the hyperparameter settings that made models most effective at capturing the statistical properties of CHILDES data, a goal which we operationalized as finding the model with the lowest perplexity.

\subsection{Hyperparameter search} 

    \paragraph{LSTMs}
    For LSTMs we explored the following hyper-parameters via a grid search for a total of 144 models.
    \begin{enumerate}
    \item layers: 2
    \item hidden and embedding size: 200, 800
    \item batch size: 20, 80
    \item dropout rate: 0.0, 0.2, 0.4, 0.6
    \item learning rate: 5.0, 10.0, 20.0
    \item random seed: 3 per parameter combination, unique for each LSTM
    \end{enumerate}

    \noindent
    The LSTM model with the lowest perplexity on the validation set after training had 2 layers, a hidden and embedding size of 800, a batch size of 20, a dropout rate of 0.4, and a learning rate of 10.\footnote{The hyperparameters we explored for the LSTMs were those of \citet{gulordava2018colorless}, the code for which can be found at \small{\texttt{ https://github.com/ facebookresearch/colorlessgreenRNNs}}} A LSTM model with these hyperparameters has 37,620,294 parameters.

    \paragraph{Transformers} For the Transformers we performed a hyperparameter sweep over the following hyper-parameters for a total of 84 models.
    \begin{enumerate}
    \item layers: 2, 4, 8, 16
    \item context size: 50, 100, 500
    \item hidden and embedding size: 200, 800, 1600
    \item heads: 2, 4, 8, 16
    \item batch size: 20, 80, 160
    \item dropout rate: 0.0, 0.2, 0.4, 0.6
    \item learning rate: 0.5, 1.0, 5.0, 10.0, 20.0
    \item random seed: 3 per parameter combination
    \end{enumerate}

    \noindent
    The Transformer model with the lowest perplexities after training had 4 layers, a context size of 500, a hidden size of 800, a batch size of 10, 4 heads, a dropout rate of 0.2, and a learning rate of 5.0. A Transformer model with these parameters has 42,759,494 parameters.

    We did not include a warmup period in our training procedure. In informal experiments, we  tried including a warmup period for both LSTMs and Transformers, but we found that this did not meaningfully affect the perplexity of the trained models in our setting.

    \subsection{Comment on model size}
    
    Although neural networks generally perform better as they increase in size, the best-performing models that we found were not the largest ones. This result is consistent with the finding of \citet{warstadt2020learning} that, for small training sets, smaller language models sometimes outperform larger ones. 
    Thus, it is unlikely that scaling up models beyond the range we investigated would have yielded better CHILDES language models than the ones we trained.

    \subsection{Implementation} 
    All models were implemented in PyTorch by building on code from \url{https://github.com/facebookresearch/colorlessgreenRNNs} and \url{https://github.com/pytorch/examples/tree/main/word\_language\_model}, and trained using Nvidia k80 GPUs. The final models are included in our GitHub repository. These models are not intended for commercial use. 
    
    \section{\textsc{Prepose-One\&Delete-One} Full Results}
    
    See Table \ref{tab:lstm-prepose-one-and-delete-one} and Table \ref{tab:transformer-prepose-one-and-delete-one} for these results.

    \begin{table}[h]
    \begin{tabular}{|l|l|l|}
    \hline
    LSTMs        & Prepose First & Prepose Main \\\hline
    Delete First & 0.01       & 0.14      \\\hline
    Delete Main  & 0.39       & 0.12      \\\hline
    Delete None  & 0.20       & 0.14      \\\hline
    \end{tabular}
    \caption{Numerical results for LSTMs’ preference for questions consistent with combinations of `prepose' and `delete' rules. Within each architecture, the proportion preferences across all 6 question types necessarily sum to 1.}
    \label{tab:lstm-prepose-one-and-delete-one}
    \end{table}
    \begin{table}[h]
    \begin{tabular}{|l|l|l|}
    \hline
    Transformers & Prepose First & Prepose Main \\\hline
    Delete First & 0.01       & 0.16      \\\hline
    Delete Main  & 0.31       & 0.06      \\\hline
    Delete None  & 0.25       & 0.21      \\\hline
    \end{tabular}
    \caption{Numerical results for Transformers’ preference for questions consistent with combinations of `prepose' and `delete' rules. Within each architecture, the proportion preferences across all 6 question types necessarily sum to 1.}
    \label{tab:transformer-prepose-one-and-delete-one}
    \end{table}

    \subsection{Results using SLOR}\label{app:slor}
    See Table \ref{lstm-cn-breakdown-slor} and Table \ref{transformer-cn-breakdown-slor} for these results. 

    \begin{table}[ht!]
    \begin{tabular}{|l|l|l|}
    \hline
    LSTMs        & Prepose First & Prepose Main  \\ \hline
    Delete First & 0.01           & 0.14          \\ \hline
    Delete Main  & 0.33          & 0.08          \\ \hline
    Delete None  & 0.26          & 0.18          \\ \hline
    \end{tabular}
    \caption{Analysis of LSTMs' preference for questions consistent with combinations of `prepose' and `delete' rules, evaluated using SLOR.  Within each architecture, the proportion preferences across all 6 question types necessarily sum to 1.} 
    \label{lstm-cn-breakdown-slor}
    \end{table} 
    
    \begin{table}[ht!]
    \begin{tabular}{|l|l|l|}
    \hline
    Transformers    & Prepose First & Prepose Main \\ \hline
    Delete First    & 0.01          & 0.15         \\ \hline
    Delete Main     & 0.27          & 0.04          \\ \hline
    Delete None     & 0.29          & 0.24         \\ \hline
    \end{tabular}
    \caption{Analysis of Transformers' preference for questions consistent with combinations of `prepose' and `delete' rules, evaluated using SLOR.  Within each architecture, the proportion preferences across all 6 question types necessarily sum to 1.} 
    \label{transformer-cn-breakdown-slor}
    \end{table}
    
    \section{BabyBERTa dataset evaluation} \label{app:babyberta}
    
\begin{figure}[t!]
        \begin{centering}
        \includegraphics[width=\columnwidth]{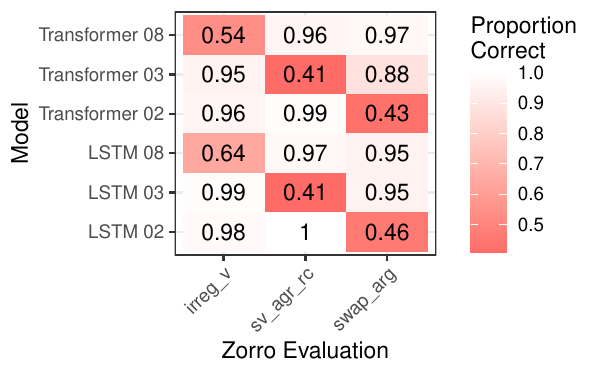}
        \caption{The performance of a selected subset of model re-runs on a selected subset of the Zorro evaluations. Each Zorro evaluation targets a specific syntactic phenomenon---in the cases shown here, irregular verbs, subject-verb agreement across relative clauses, and correct argument ordering.} 
        \label{fig:babyberta-subset-heatmap}
        \end{centering}
    \end{figure}

\begin{figure*}[ht!]
        \begin{centering}
        \includegraphics[width=\textwidth]{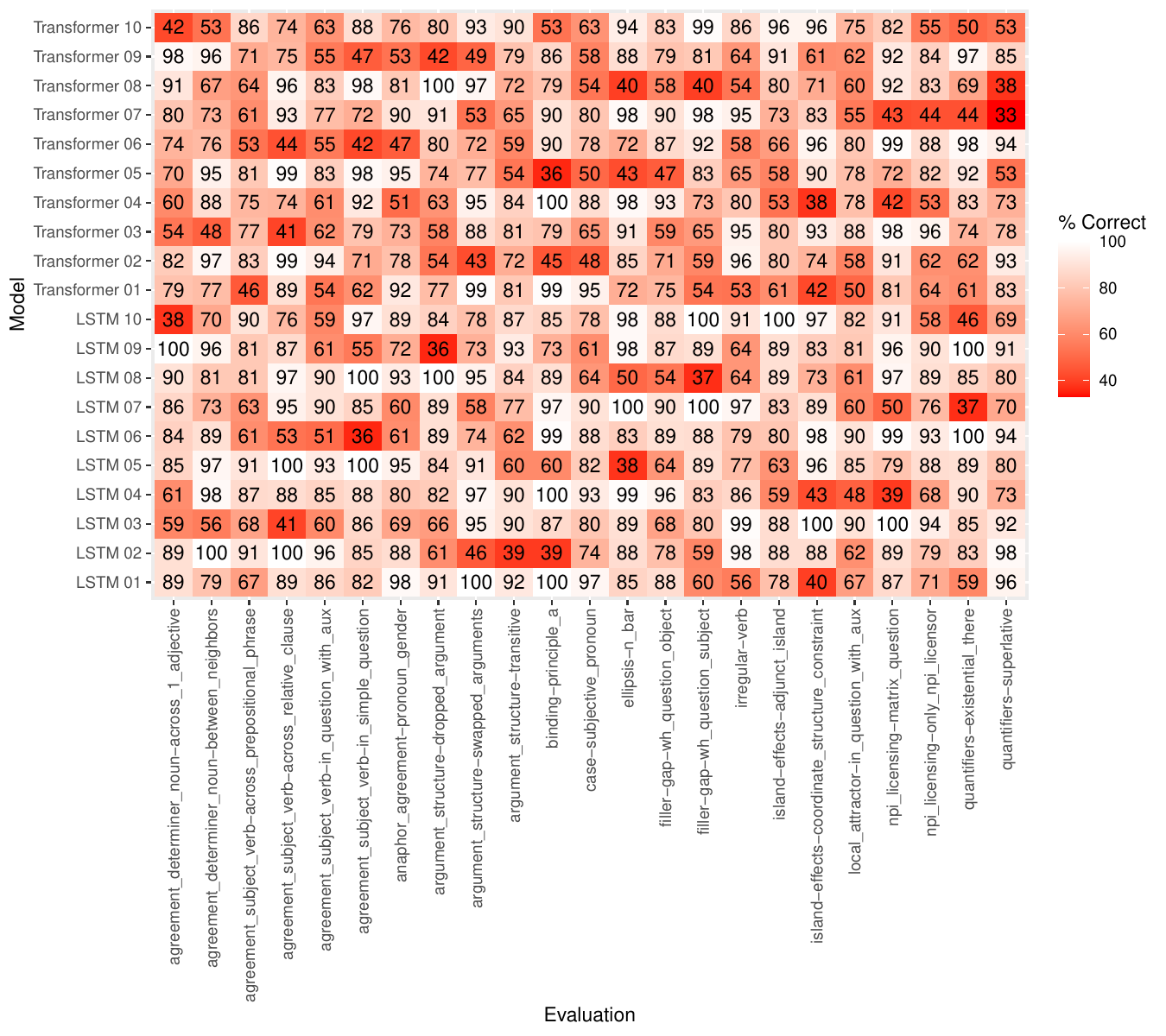}
        \caption{Results on the targeted syntactic evaluations in \citet{huebner2021babyberta} in percent accuracy.  Evaluation names in Figure \ref{fig:babyberta-subset-heatmap} were shortened.} 
        \label{fig:babyberta-heatmap}
        \end{centering}
    \end{figure*}

    For an illustrative subset of the results on the Zorro evaluation dataset (discussed in Section \ref{sec:zorro}), see Figure \ref{fig:babyberta-subset-heatmap}. For the full results, see Figure \ref{fig:babyberta-heatmap}.

    \section{Move-One Dataset Results}
    One approach used in several past papers (e.g., \citet{lewis2001learnability} and \citet{reali2005uncovering}) is to evaluate models using pairs of sentences that can be formed by starting with a declarative sentence (e.g., \ref{ex:2fc_origin}) and moving one of its auxiliaries to the front of the sentence. The first sentence in each pair (e.g., \ref{ex:2fc_hierarchical_q} ) follows \textsc{HierarchicalQ}, because the \textit{main} auxiliary is moved, while the second (e.g., \ref{ex:2fc_linear_q}), follows \textsc{LinearQ} because the \textit{first} auxiliary is moved.

     \ex. The children who \textcolor{blue}{\textbf{are}} talking \textcolor{blue}{\textbf{are}} sleeping.\label{ex:2fc_origin}
     
     \ex. \label{ex:2fc}
     \a. \textcolor{blue}{\textbf{Are}} the children who \textcolor{blue}{\textbf{are}} talking sleeping?\label{ex:2fc_hierarchical_q}
     \b. \textcolor{blue}{\textbf{Are}} the children who talking \textcolor{blue}{\textbf{are}} sleeping?\label{ex:2fc_linear_q}
     
     \noindent
     If a model assigns a higher probability to \ref{ex:2fc_hierarchical_q} than \ref{ex:2fc_linear_q}, that is evidence that the models favors \textsc{HierarchicalQ} over \textsc{LinearQ}. While this preference is a necessary component of correctly learning \textsc{HierarchicalQ}, it is by no means sufficient:
     indeed, \citet{kam2008} showed that models can prefer sentences consistent with \textsc{HierarchicalQ} over sentences consistent with \textsc{LinearQ} due to shallow $n$-gram statistics rather than due to knowledge of hierarchical structure.
     More generally, there are infinitely many other incorrect hypotheses besides \textsc{LinearQ}, and demonstrating successful learning of \textsc{HierarchicalQ} would require ruling out all of them. Investigating all possibilities is intractable, but we can at least investigate a few additional plausible ones. Thus, in the main paper we depart from prior work by considering a greater number of candidate sentences than just the pairs of sentences used in prior work.

     To create the \textsc{Move-One} dataset, we randomly sampled 10,000 declarative sentences from our CFGs for which the first and main auxiliary were identical and then modified them to give 10,000 sentence pairs. To create the \textsc{Prepose-One\&Delete-One} dataset, we randomly  sampled a different 10,000 declarative sentences from our CFGs for which the first and main auxiliary were different and then we modified them to give 10,000 6-tuples of sentences. See Appendix \ref{app:cfgs} for more details about the CFGs.

    \section{Context Free Grammars}\label{app:cfgs}
    Figure~\ref{cfg-prepose-one-and-delete-one} contains the context-free grammar used for the analyses in Section~\ref{sec:yn_questions_results}. 
    Figures \ref{cfg-first-aux-equal-to-main-aux} and \ref{cfg-first-aux-not-equal-to-main-aux} contain the context-free grammars used for the targeted evaluation sets in Section~\ref{sec:question_formation_results}; for each of these evaluation sets, we sampled 10,000 declarative sentences from these grammars and transformed them into questions according to \textsc{HierarchicalQ}. Figure~\ref{cfg-vocabulary} contains the vocabulary used for all of these datasets.

\begin{figure*}[ht!]
\begin{center}
\begin{tabular}{p{2cm}p{12cm}}
\toprule
S &$\rightarrow$ \{NP\_S RC\_S\_BARE MAIN-AUX VP\_S\_PAST\}\\
S &$\rightarrow$ \{NP\_S RC\_S\_PAST MAIN-AUX VP\_S\_BARE\}\\
S &$\rightarrow$ \{NP\_S RC\_S\_BARE MAIN-AUX VP\_S\_PROG\}\\
S &$\rightarrow$ \{NP\_S RC\_S\_PROG MAIN-AUX VP\_S\_BARE\}\\
S &$\rightarrow$ \{NP\_S RC\_S\_PAST MAIN-AUX VP\_S\_PROG\}\\
S &$\rightarrow$ \{NP\_S RC\_S\_PROG MAIN-AUX VP\_S\_PAST\}\\

S &$\rightarrow$ \{NP\_P RC\_P\_BARE MAIN-AUX VP\_P\_PAST\}\\
S &$\rightarrow$ \{NP\_P RC\_P\_PAST MAIN-AUX VP\_P\_BARE\}\\
S &$\rightarrow$ \{NP\_P RC\_P\_BARE MAIN-AUX VP\_P\_PROG\}\\
S &$\rightarrow$ \{NP\_P RC\_P\_PROG MAIN-AUX VP\_P\_BARE\}\\
S &$\rightarrow$ \{NP\_P RC\_P\_PAST MAIN-AUX VP\_P\_PROG\}\\
S &$\rightarrow$ \{NP\_P RC\_P\_PROG MAIN-AUX VP\_P\_PAST\}\\

NP\_S &$\rightarrow$ \{Det\_S N\_S\}\\
NP\_P &$\rightarrow$ \{Det\_P N\_P\}\\

NP\_O &$\rightarrow$ \{Det\_S N\_S $|$ Det\_P N\_P $|$ Det\_S N\_S Prep Det\_S N\_S $|$ Det\_S N\_S Prep Det\_P N\_P $|$ Det\_P N\_P Prep Det\_S N\_S $|$ Det\_P N\_P Prep Det\_P N\_P\}\\

VP\_S\_BARE &$\rightarrow$ \{Aux\_S IV \}\\
VP\_S\_BARE &$\rightarrow$ \{Aux\_S TV NP\_O\}\\
VP\_S\_PROG &$\rightarrow$ \{Aux\_S\_BE IV\_IS\}\\
VP\_S\_PROG &$\rightarrow$ \{Aux\_S\_BE TV\_IS NP\_O\}\\
VP\_S\_PAST &$\rightarrow$ \{Aux\_S\_HAS IV\_HAS\}\\
VP\_S\_PAST &$\rightarrow$ \{Aux\_S\_HAS TV\_HAS NP\_O\}\\

VP\_P\_BARE &$\rightarrow$ \{Aux\_P IV\}\\
VP\_P\_BARE &$\rightarrow$ \{Aux\_P TV NP\_O\}\\
VP\_P\_PROG &$\rightarrow$ \{Aux\_P\_BE IV\_IS\}\\
VP\_P\_PROG &$\rightarrow$ \{Aux\_P\_BE TV\_IS NP\_O\}\\
VP\_P\_PAST &$\rightarrow$ \{Aux\_P\_HAS IV\_HAS\}\\
VP\_P\_PAST &$\rightarrow$ \{Aux\_P\_HAS TV\_HAS NP\_O\}\\

RC\_S\_BARE &$\rightarrow$ \{Rel Aux\_S IV $|$ Rel Det\_S N\_S Aux\_S TV $|$ Rel Det\_P N\_P Aux\_P TV $|$ Rel Aux\_S TV Det\_S N\_S $|$ Rel Aux\_S TV Det\_P N\_P\}\\
RC\_S\_PROG &$\rightarrow$ \{Rel Aux\_S\_BE IV\_IS $|$ Rel Det\_S N\_S Aux\_S\_BE TV\_IS $|$ Rel Det\_P N\_P Aux\_P\_BE TV\_IS $|$ Rel Aux\_S\_BE TV\_IS Det\_S N\_S $|$ Rel Aux\_S\_BE TV\_IS Det\_P N\_P\}\\
RC\_S\_PAST &$\rightarrow$ \{Rel Aux\_S\_HAS IV\_HAS $|$ Rel Det\_S N\_S Aux\_S\_HAS TV\_HAS $|$ Rel Det\_P N\_P Aux\_P\_HAS TV\_HAS $|$ Rel Aux\_S\_HAS TV\_HAS Det\_S N\_S $|$ Rel Aux\_S\_HAS TV\_HAS Det\_P N\_P\}\\
RC\_P\_BARE &$\rightarrow$ \{Rel Aux\_P IV $|$ Rel Det\_S N\_S Aux\_S TV $|$ Rel Det\_P N\_P Aux\_P TV $|$ Rel Aux\_P TV Det\_S N\_S $|$ Rel Aux\_P TV Det\_P N\_P\}\\
RC\_P\_PROG &$\rightarrow$ \{Rel Aux\_P\_BE IV\_IS $|$ Rel Det\_S N\_S Aux\_S\_BE TV\_IS $|$ Rel Det\_P N\_P Aux\_P\_BE TV\_IS $|$ Rel Aux\_P\_BE TV\_IS Det\_S N\_S $|$ Rel Aux\_P\_BE TV\_IS Det\_P N\_P\}\\
RC\_P\_PAST &$\rightarrow$ \{Rel Aux\_P\_HAS IV\_HAS $|$ Rel Det\_S N\_S Aux\_S\_HAS TV\_HAS $|$ Rel Det\_P N\_P Aux\_P\_HAS TV\_HAS $|$ Rel Aux\_P\_HAS TV\_HAS Det\_S N\_S $|$ Rel Aux\_P\_HAS TV\_HAS Det\_P N\_P\}\\
\bottomrule
\end{tabular}
\end{center}
\caption{CFG used to generate \textsc{Prepose-One-And-Delete-One} evaluation dataset}
\label{cfg-prepose-one-and-delete-one}
\end{figure*}

\begin{figure*}[ht!]
\begin{center}
\begin{tabular}{p{1cm}p{12cm}}
\toprule
S &$\rightarrow$ \{NP\_M\_S VP\_M\_S $|$ NP\_M\_P VP\_M\_P\}\\
NP\_M\_S &$\rightarrow$ \{Det\_S N\_S $|$ Det\_S N\_S Prep Det\_S N\_S $|$ Det\_S N\_S Prep Det\_P N\_P\}\\
NP\_M\_P &$\rightarrow$ \{Det\_P N\_P $|$ Det\_P N\_P Prep Det\_S N\_S $|$ Det\_P N\_P Prep Det\_P N\_P\}\\
NP\_O &$\rightarrow$ \{Det\_S N\_S $|$ Det\_P N\_P $|$ Det\_S N\_S Prep Det\_S N\_S $|$ Det\_S N\_S Prep Det\_P N\_P $|$ Det\_P N\_P Prep Det\_S N\_S $|$  Det\_P N\_P Prep Det\_P N\_P $|$ Det\_S N\_S RC\_S $|$ Det\_P N\_P RC\_P \}\\
VP\_M\_S &$\rightarrow$ \{Aux\_S IV \}\\
VP\_M\_S &$\rightarrow$ \{Aux\_S TV NP\_O\}\\
VP\_M\_S &$\rightarrow$ \{Aux\_S\_BE IV\_IS\}\\
VP\_M\_S &$\rightarrow$ \{Aux\_S\_BE TV\_IS NP\_O\}\\
VP\_M\_S &$\rightarrow$ \{Aux\_S\_HAS IV\_HAS\}\\
VP\_M\_S &$\rightarrow$ \{Aux\_S\_HAS TV\_HAS NP\_O\}\\
VP\_M\_P &$\rightarrow$ \{Aux\_P IV\}\\
VP\_M\_P &$\rightarrow$ \{Aux\_P TV NP\_O\}\\
VP\_M\_P &$\rightarrow$ \{Aux\_P\_BE IV\_IS\}\\
VP\_M\_P &$\rightarrow$ \{Aux\_P\_BE TV\_IS NP\_O\}\\
VP\_M\_P &$\rightarrow$ \{Aux\_P\_HAS IV\_HAS\}\\
VP\_M\_P &$\rightarrow$ \{Aux\_P\_HAS TV\_HAS NP\_O\}\\
RC\_S &$\rightarrow$ \{Rel Aux\_S IV $|$ Rel Det\_S N\_S Aux\_S TV $|$ Rel Det\_P N\_P Aux\_P TV $|$ Rel Aux\_S TV Det\_S N\_S $|$ Rel Aux\_S TV Det\_P N\_P\}\\
RC\_S &$\rightarrow$ \{Rel Aux\_S\_BE IV\_IS $|$ Rel Det\_S N\_S Aux\_S\_BE TV\_IS $|$ Rel Det\_P N\_P Aux\_P\_BE TV\_IS $|$ Rel Aux\_S\_BE TV\_IS Det\_S N\_S $|$ Rel Aux\_S\_BE TV\_IS Det\_P N\_P\}\\
RC\_S &$\rightarrow$ \{Rel Aux\_S\_HAS IV\_HAS $|$ Rel Det\_S N\_S Aux\_S\_HAS TV\_HAS $|$ Rel Det\_P N\_P Aux\_P\_HAS TV\_HAS $|$ Rel Aux\_S\_HAS TV\_HAS Det\_S N\_S $|$ Rel Aux\_S\_HAS TV\_HAS Det\_P N\_P\}\\
RC\_P &$\rightarrow$ \{Rel Aux\_P IV $|$ Rel Det\_S N\_S Aux\_S TV $|$ Rel Det\_P N\_P Aux\_P TV $|$ Rel Aux\_P TV Det\_S N\_S $|$ Rel Aux\_P TV Det\_P N\_P\}\\
RC\_P &$\rightarrow$ \{Rel Aux\_P\_BE IV\_IS $|$ Rel Det\_S N\_S Aux\_S\_BE TV\_IS $|$ Rel Det\_P N\_P Aux\_P\_BE TV\_IS $|$ Rel Aux\_P\_BE TV\_IS Det\_S N\_S $|$ Rel Aux\_P\_BE TV\_IS Det\_P N\_P\}\\
RC\_P &$\rightarrow$ \{Rel Aux\_P\_HAS IV\_HAS $|$ Rel Det\_S N\_S Aux\_S\_HAS TV\_HAS $|$ Rel Det\_P N\_P Aux\_P\_HAS TV\_HAS $|$ Rel Aux\_P\_HAS TV\_HAS Det\_S N\_S $|$ Rel Aux\_P\_HAS TV\_HAS Det\_P N\_P\}\\
\bottomrule
\end{tabular}
\end{center}
\caption{CFG used to generate \textsc{First-Aux = Main-Aux} evaluation dataset}
\label{cfg-first-aux-equal-to-main-aux}
\end{figure*}

\begin{figure*}[ht!]
\begin{center}
\begin{tabular}{p{1cm}p{12cm}}
\toprule
S &$\rightarrow$ \{NP\_M\_S VP\_M\_S $|$ NP\_M\_P VP\_M\_P\}\\
NP\_M\_S &$\rightarrow$ \{Det\_S N\_S $|$ Det\_S N\_S Prep Det\_S N\_S $|$ Det\_S N\_S Prep Det\_P N\_P\}\\
NP\_M\_P &$\rightarrow$ \{Det\_P N\_P $|$ Det\_P N\_P Prep Det\_S N\_S $|$ Det\_P N\_P Prep Det\_P N\_P\}\\
NP\_O &$\rightarrow$ \{Det\_S N\_S $|$ Det\_P N\_P $|$ Det\_S N\_S Prep Det\_S N\_S $|$ Det\_S N\_S Prep Det\_P N\_P $|$ Det\_P N\_P Prep Det\_S N\_S $|$  Det\_P N\_P Prep Det\_P N\_P $|$ Det\_S N\_S RC\_S $|$ Det\_P N\_P RC\_P \}\\
VP\_M\_S &$\rightarrow$ \{Aux\_S IV \}\\
VP\_M\_S &$\rightarrow$ \{Aux\_S TV NP\_O\}\\
VP\_M\_S &$\rightarrow$ \{Aux\_S\_BE IV\_IS\}\\
VP\_M\_S &$\rightarrow$ \{Aux\_S\_BE TV\_IS NP\_O\}\\
VP\_M\_S &$\rightarrow$ \{Aux\_S\_HAS IV\_HAS\}\\
VP\_M\_S &$\rightarrow$ \{Aux\_S\_HAS TV\_HAS NP\_O\}\\
VP\_M\_P &$\rightarrow$ \{Aux\_P IV\}\\
VP\_M\_P &$\rightarrow$ \{Aux\_P TV NP\_O\}\\
VP\_M\_P &$\rightarrow$ \{Aux\_P\_BE IV\_IS\}\\
VP\_M\_P &$\rightarrow$ \{Aux\_P\_BE TV\_IS NP\_O\}\\
VP\_M\_P &$\rightarrow$ \{Aux\_P\_HAS IV\_HAS\}\\
VP\_M\_P &$\rightarrow$ \{Aux\_P\_HAS TV\_HAS NP\_O\}\\
RC\_S &$\rightarrow$ \{Rel Aux\_S IV $|$ Rel Det\_S N\_S Aux\_S TV $|$ Rel Det\_P N\_P Aux\_P TV $|$ Rel Aux\_S TV Det\_S N\_S $|$ Rel Aux\_S TV Det\_P N\_P\}\\
RC\_S &$\rightarrow$ \{Rel Aux\_S\_BE IV\_IS $|$ Rel Det\_S N\_S Aux\_S\_BE TV\_IS $|$ Rel Det\_P N\_P Aux\_P\_BE TV\_IS $|$ Rel Aux\_S\_BE TV\_IS Det\_S N\_S $|$ Rel Aux\_S\_BE TV\_IS Det\_P N\_P\}\\
RC\_S &$\rightarrow$ \{Rel Aux\_S\_HAS IV\_HAS $|$ Rel Det\_S N\_S Aux\_S\_HAS TV\_HAS $|$ Rel Det\_P N\_P Aux\_P\_HAS TV\_HAS $|$ Rel Aux\_S\_HAS TV\_HAS Det\_S N\_S $|$ Rel Aux\_S\_HAS TV\_HAS Det\_P N\_P\}\\
RC\_P &$\rightarrow$ \{Rel Aux\_P IV $|$ Rel Det\_S N\_S Aux\_S TV $|$ Rel Det\_P N\_P Aux\_P TV $|$ Rel Aux\_P TV Det\_S N\_S $|$ Rel Aux\_P TV Det\_P N\_P\}\\
RC\_P &$\rightarrow$ \{Rel Aux\_P\_BE IV\_IS $|$ Rel Det\_S N\_S Aux\_S\_BE TV\_IS $|$ Rel Det\_P N\_P Aux\_P\_BE TV\_IS $|$ Rel Aux\_P\_BE TV\_IS Det\_S N\_S $|$ Rel Aux\_P\_BE TV\_IS Det\_P N\_P\}\\
RC\_P &$\rightarrow$ \{Rel Aux\_P\_HAS IV\_HAS $|$ Rel Det\_S N\_S Aux\_S\_HAS TV\_HAS $|$ Rel Det\_P N\_P Aux\_P\_HAS TV\_HAS $|$ Rel Aux\_P\_HAS TV\_HAS Det\_S N\_S $|$ Rel Aux\_P\_HAS TV\_HAS Det\_P N\_P\}\\

\bottomrule
\end{tabular}
\end{center}
\caption{CFG used to generate \textsc{First-Aux $\not=$ Main-Aux} evaluation dataset}
\label{cfg-first-aux-not-equal-to-main-aux}
\end{figure*}

\begin{figure*}[ht!]
\begin{center}
\begin{tabular}{p{1.5cm}p{12cm}}
\toprule
Det\_S &$\rightarrow$ \{the $|$ some $|$ this \}\\
Det\_P &$\rightarrow$ \{the $|$ some $|$ those\}\\
N\_S &$\rightarrow$ \{baby $|$ girl $|$ boy $|$ animal $|$ child $|$ person $|$ horse \}\\
N\_P &$\rightarrow$ \{babies $|$ girls $|$ boys $|$ animals $|$ children $|$ people $|$ horses \}\\
IV &$\rightarrow$ \{play $|$ read $|$ draw $|$ sit $|$ fall $|$ talk $|$ sleep $|$ try $|$ work $|$ walk\}\\
IV\_IS &$\rightarrow$ \{playing $|$ reading $|$ drawing $|$ sitting $|$ falling $|$ talking $|$ sleeping $|$ trying $|$ working $|$ walking\}\\
IV\_HAS &$\rightarrow$ \{played $|$ read $|$ drawn $|$ sat $|$ fallen $|$ talked $|$ slept $|$ tried $|$ worked $|$ walked\}\\
TV &$\rightarrow$ \{call $|$ see $|$ find $|$ help $|$ feed $|$ know $|$ pick $|$ visit $|$ watch $|$ reach\}\\
TV\_IS &$\rightarrow$ \{calling $|$ seeing $|$ finding $|$ helping $|$ feeding $|$ knowing $|$ picking $|$ visiting $|$ watching $|$ reaching\}
\\
TV\_HAS &$\rightarrow$ \{called $|$ seen $|$ found $|$ helped $|$ fed $|$ known $|$ picked $|$ visited $|$ watched $|$ reached\}\\
Aux\_P &$\rightarrow$ \{do $|$ did $|$ can $|$ would $|$ shall\}\\
Aux\_S &$\rightarrow$ \{does $|$ did $|$ can $|$ would $|$ shall\}\\
Aux\_S\_BE &$\rightarrow$ \{is $|$ was\}\\
Aux\_P\_BE &$\rightarrow$ \{are $|$ were\}\\
Aux\_S\_HAS &$\rightarrow$ \{has\}\\
Aux\_P\_HAS &$\rightarrow$ \{have\}\\
Prep &$\rightarrow$ \{by $|$ behind \}\\
Rel &$\rightarrow$ \{who $|$ that \}\\
\bottomrule
\end{tabular}
\end{center}
\caption{Vocabulary used for the \textsc{Prepose-One-And-Delete-One}, \textsc{First-Aux $\not =$ Main-Aux}, and \textsc{First-Aux = Main-Aux} evaluation datasets}
\label{cfg-vocabulary}
\end{figure*}

    \section{Breakdown by lexical identity}\label{app:lex_id}
    
\begin{figure*}[ht!]
        \centering
        \includegraphics[width=400px]{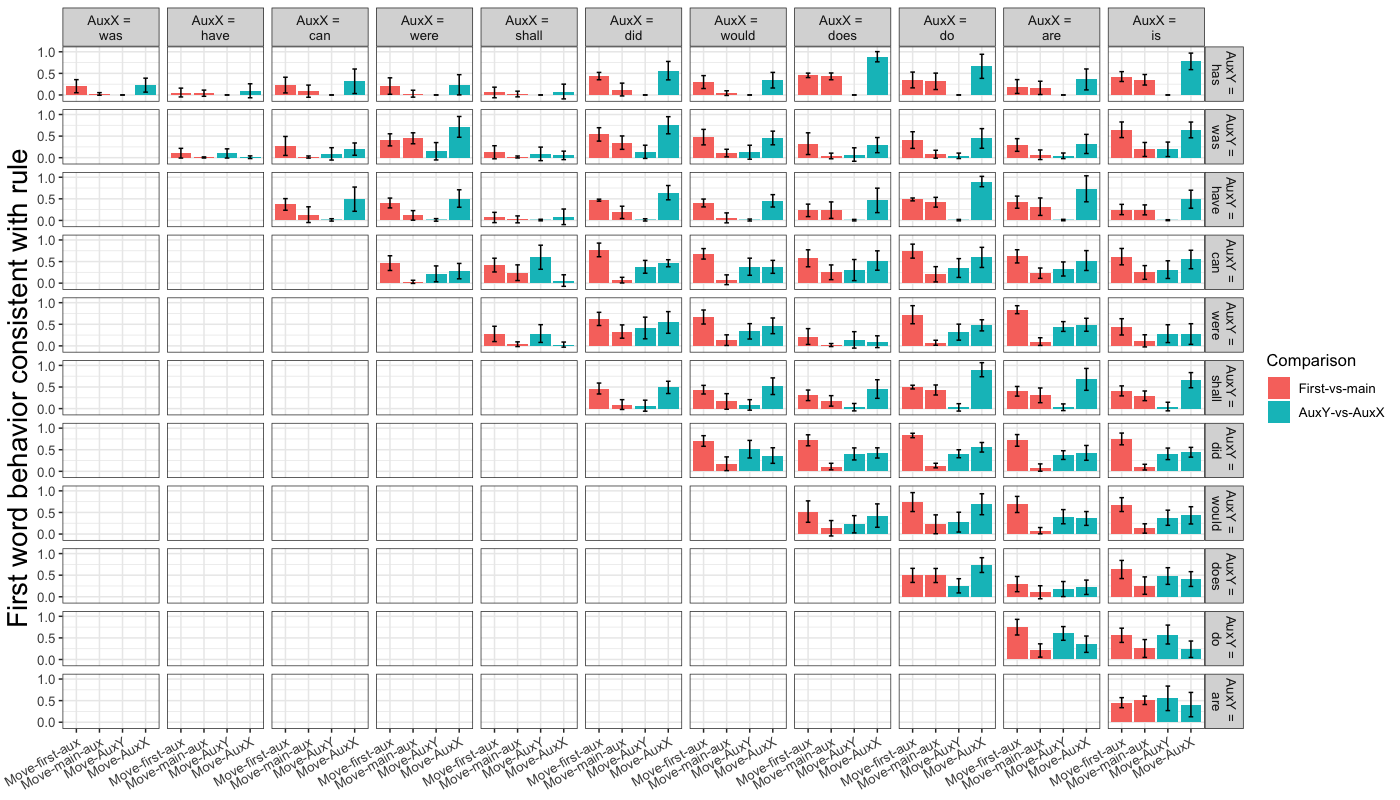}
        \caption{Breakdown by the identities of the two auxiliaries for outputs in the \textsc{First-Aux $\neq$ Main-Aux} evaluation set for LSTMs trained only on question formation. The two leftmost bars in each cell show a First-vs-main comparison, while the two rightmost bars show an AuxY-vs-AuxX comparison.}
        \label{lexicallly-specific-LSTM-pretrained}
    \end{figure*}

\begin{figure*}[ht!]
        \centering
        \includegraphics[width=400px]{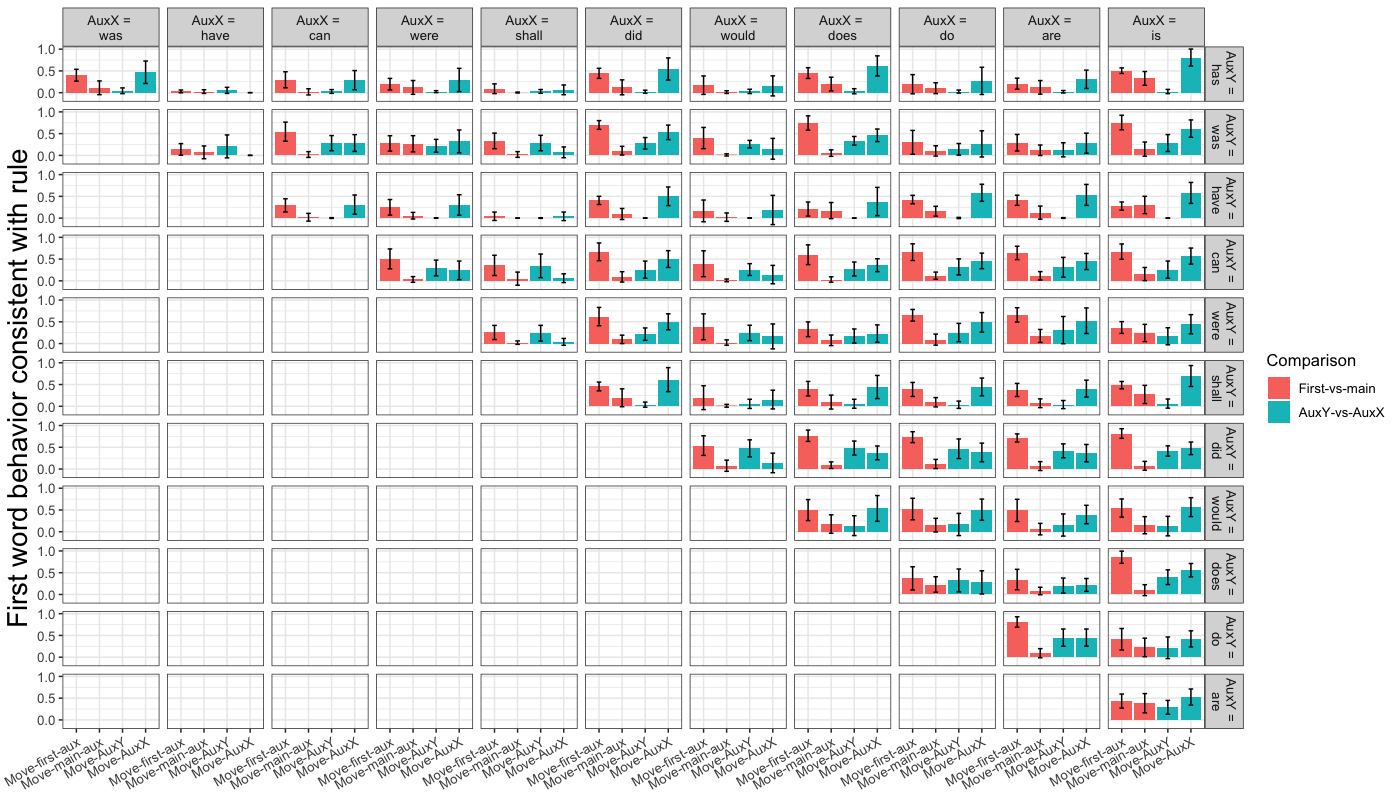}
        \caption{Breakdown by the identities of the two auxiliaries for outputs in the \textsc{First-Aux $\neq$ Main-Aux} evaluation set for LSTMs first trained on next-word prediction and then question formation. The two leftmost bars in each cell show a First-vs-main comparison, while the two rightmost bars show an AuxY-vs-AuxX comparison.}
        \label{lexically-specific-LSTM-finetuned}
    \end{figure*}

\begin{figure*}[ht!]
        \centering
        \includegraphics[width=400px]{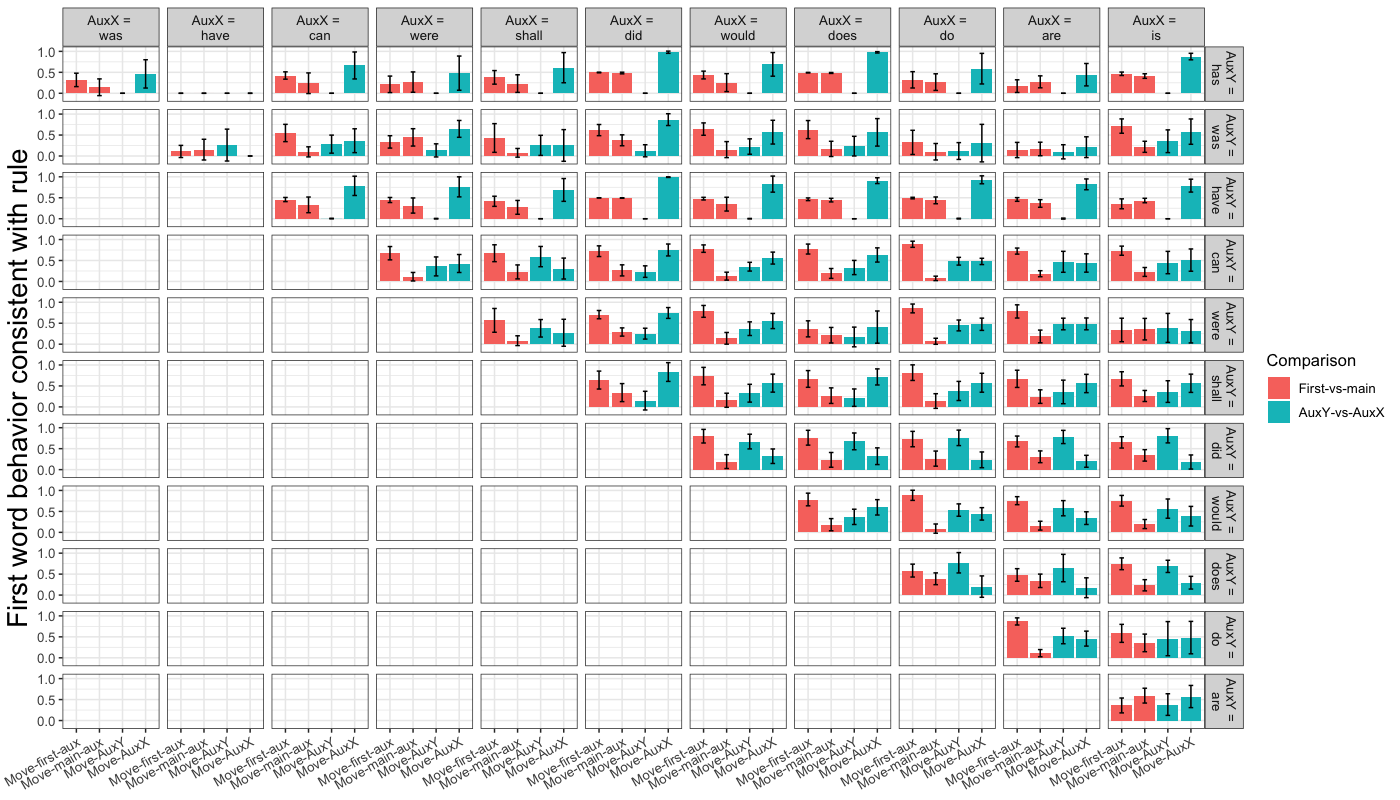}
        \caption{Breakdown by the identities of the two auxiliaries for outputs in the \textsc{First-Aux $\neq$ Main-Aux} evaluation set for Transformers trained only on question formation. The two leftmost bars in each cell show a First-vs-main comparison, while the two rightmost bars show an AuxY-vs-AuxX comparison.}
        \label{lexically-specific-Transformer-pretrained}
    \end{figure*}

\begin{figure*}[ht!]
        \centering
        \includegraphics[width=400px]{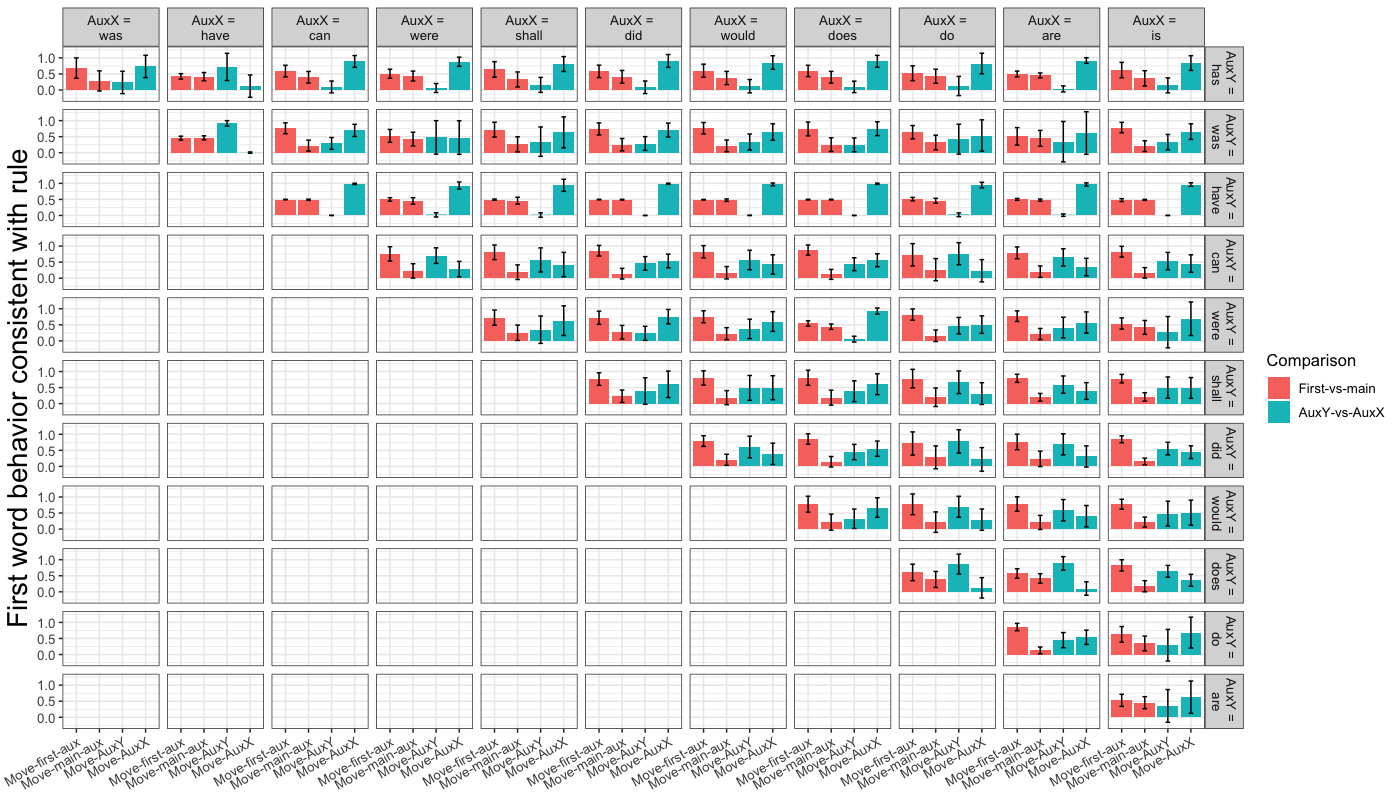}
        \caption{Breakdown by the identities of the two auxiliaries for outputs in the \textsc{First-Aux $\neq$ Main-Aux} evaluation set for Transformers first trained on next-word prediction and then question formation. The two leftmost bars in each cell show a First-vs-main comparison, while the two rightmost bars show an AuxY-vs-AuxX comparison.}
        \label{lexically-specific-Transformer-finetuned}
    \end{figure*}

    Here we further break down models' predictions for the \textsc{First-Aux $\neq$ Main-Aux} evaluation set based on the identities of the two auxiliaries in the input sentence.
    Figure \ref{lexicallly-specific-LSTM-pretrained} gives the results for the LSTM in the \textsc{Question Formation} condition; Figure \ref{lexically-specific-LSTM-finetuned} for the LSTM in the \textsc{Next-Word Prediction + Question Formation} condition; Figure \ref{lexically-specific-Transformer-pretrained} for the Transformer in the \textsc{Question Formation} condition; and Figure \ref{lexically-specific-Transformer-finetuned} for the for the Transformer in the \textsc{Next-Word Prediction + Question Formation} condition.
    

    \section{Example generated text}\label{app:generated_text}
    
    Figure \ref{generated-text} gives some example text generated by our models.
    Models trained on next-word prediction produce their predictions as a probability distribution over the vocabulary. To use such models to generate text, we sample a word from this distribution then use that word as the model's input for the next time step.
           
        \begin{figure*}[ht!]
        \begin{center}
        \begin{tabular}{p{2cm}p{12cm}}
        \toprule
            \textbf{Text from CHILDES} & do you eat in nursery school too ? oh . it 's nursery school . xxx . what do you get to eat ? what else do you get to eat besides a book ? do n't play with those books . what else do you have ? what do you have in nursery school ? alright - you can take it . why do n't you put the turtle back . well - you 're not going to change his water now . tonight . tonight - you change it about ten times a day . no - you can do it later . forget it now . too bad . too bad . that poor turtle . you do n't let him sit there five minutes . that was five minutes ?\\
            \textbf{LSTM generated text} & \underline{do you eat in nursery school too ? oh . it 's nursery school .} no ! tomorrow 's Friday . right . yeah - yeah . yeah I know what . I told you that he liked snipsnap that way . well this is address the grayish kings . ok . what kind of bird do you want to play with ? what else do you want to do ? your crayons . got a circle . I 'm going to use these . there . remember ? ok do you want to put all of these there ? ok let 's put everything away . what do you want to do ? you want to play in the box ? it 's just like a ball because it has a ball on it .\\
            \textbf{Transformer generated text} & \underline{do you eat in nursery school too ? oh . it 's nursery school .}  mm-hm . and it looks like a little girl . mm-hm . like a little girl . hm , mm-hm . looks like she 's concerned about baseball this is a picture of a double football player . mm-hm . um , it has a uniform on it . mm-hm . anything more ? no . ok great . got a hat on . and a flower and a little flower . remember that ? do you remember that word ? what do you remember that word ? do you remember what it word says ? it says seven - eight . what 's that word ? it says eight . look at that word . three - four - six ... eight - nine ...
            \\ \bottomrule
            \end{tabular}
            \end{center}
            \caption{Comparison of text generated by the LSTM and Transformer models with a block of text chosen randomly from the training data. The LSTMs and Transformers were both prompted with the first three sentences of the text taken from CHILDES, which is the underlined in the two model generated texts. Note that neither of the model generated texts were cherry-picked either for quality or to be representative of the models' usual output: rather they were the first things they generated when prompted with the above underlined portion.}
            \label{generated-text}
        \end{figure*}

\end{document}